\documentclass[conference]{IEEEtran}
\IEEEoverridecommandlockouts

\usepackage[style=ieee,maxcitenames=2,mincitenames=1]{biblatex}
\AtEveryBibitem{%
  \clearfield{note}%
}

\usepackage{amsmath,amssymb,amsfonts}
\usepackage{graphicx}
\usepackage{textcomp}
\usepackage{xcolor}
\def\BibTeX{{\rm B\kern-.05em{\sc i\kern-.025em b}\kern-.08em
    T\kern-.1667em\lower.7ex\hbox{E}\kern-.125emX}}

\usepackage{hyperref}
\hypersetup{
  colorlinks,
  citecolor= [rgb]{0, 0.720, 0.051},
  linkcolor= [rgb]{0.847, 0.106, 0.376},
  urlcolor= [rgb]{0.118, 0.533, 0.898}
}
\usepackage{algorithm}
\usepackage{algpseudocode}

\usepackage{url}
\usepackage{booktabs}
\usepackage{multirow}
\usepackage{soul}
\usepackage{stfloats, caption}%
\usepackage{lipsum}
\usepackage{xspace}
\usepackage{subcaption}

\usepackage[most]{tcolorbox}

\tcbset{textmarker/.style={%
        parbox=true,boxrule=0mm,boxsep=0mm,arc=0mm,
        outer arc=0mm,left=1mm,right=1mm,top=7pt,bottom=7pt,
        toptitle=1mm,bottomtitle=1mm,oversize}}

\newtcolorbox{highlightBox}{textmarker,
    colback=yellow!10!white,
    colframe=yellow!40!black,
    fonttitle=\bfseries,
    title=Highlights,
    sharp corners,
    boxrule=0.4mm,
    }

\newcommand{\highlight}[1]{\begin{highlightBox} #1 \end{highlightBox}}
\newcommand{\algo}{\texttt{TS-Inverse}\xspace}

\newif\ifcomm
\commtrue

\newif\ifanony
\anonyfalse

\newif\ifpreprint
\preprinttrue

\usepackage{fancyhdr}

\fancypagestyle{plain}{
  \fancyhf{}

  \fancyfoot[C]{\footnotesize This work has been accepted for publication in the 3rd IEEE Conference on Secure and Trustworthy Machine Learning (SaTML 2025). \\ The final version will be available on IEEE Xplore.}
}

\ifpreprint
\fi

\begin{document}

\title{TS-Inverse: A Gradient Inversion Attack Tailored for Federated Time Series Forecasting Models

\thanks{This research was partially supported by the SNSF project, Priv-GSyn (grant number 200021E\_229204), and the Advanced Computing Engineering (ACE) Department at TNO.}

}

\author{

\ifanony 
    \IEEEauthorblockN{Anonymous Authors}
    \IEEEauthorblockA{Anonymous Affiliations}
\else
\IEEEauthorblockN{Caspar Meijer\IEEEauthorrefmark{1}\IEEEauthorrefmark{2}}
\IEEEauthorblockA{caspar.meijer@tno.nl}
\and
\IEEEauthorblockN{Jiyue Huang\IEEEauthorrefmark{1}}
\IEEEauthorblockA{j.huang-4@tudelft.nl}
\and
\IEEEauthorblockN{Shreshtha Sharma\IEEEauthorrefmark{2}}
\IEEEauthorblockA{shreshtha.sharma@tno.nl}
\and
\IEEEauthorblockN{Elena Lazovik\IEEEauthorrefmark{2}}
\IEEEauthorblockA{elena.lazovik@tno.nl}
\and
\IEEEauthorblockN{Lydia Y. Chen\IEEEauthorrefmark{1}\IEEEauthorrefmark{3}}
\IEEEauthorblockA{lydiaychen@ieee.org}
\and
\IEEEauthorblockA{\centerline{\IEEEauthorrefmark{1}Delft University of Technology, Delft, The Netherlands}}
\IEEEauthorblockA{\centerline{\IEEEauthorrefmark{2}Dutch Organization for Applied Scientific Research (TNO), The Hague, The Netherlands}}
\IEEEauthorblockA{\centerline{\IEEEauthorrefmark{3}University of Neuchâtel, Neuchâtel, Switzerland}}
\fi
}

\maketitle

\ifpreprint
    \thispagestyle{plain}
\fi

\begin{abstract}
Federated learning (FL) for time series forecasting (TSF) enables clients with privacy-sensitive time series (TS) data to collaboratively learn accurate forecasting models, for example, in energy load prediction.
Unfortunately, privacy risks in FL persist, as servers can potentially reconstruct clients' training data through gradient inversion attacks (GIA). 
Although GIA is demonstrated for image classification tasks, little is known about time series regression tasks.
In this paper, we first conduct an extensive empirical study on inverting TS data across 4 TSF models and 4 datasets, identifying the unique challenges of reconstructing both observations and targets of TS data. 
We then propose TS-Inverse, a novel GIA that improves the inversion of TS data by (i) learning a gradient inversion model that outputs quantile predictions, (ii) a unique loss function that incorporates periodicity and trend regularization, and (iii) regularization according to the quantile predictions. Our evaluations demonstrate a remarkable performance of TS-Inverse, achieving at least a 2x-10x improvement in terms of the sMAPE metric over existing GIA methods on TS data. Code repository: 
\ifanony 
    \url{www.github.com/<anonymous>/<anonymous>}
\else
    \url{https://github.com/Capsar/ts-inverse}
\fi
\end{abstract}


\section{Introduction} \label{sec:introduction}
Federated learning (FL), a distributed machine learning framework in which a server and multiple clients collaboratively train a global model, is established as an effective method for training deep neural networks without centrally storing client training data. In this framework, clients independently train the model on their local data, and then transmit these updates to a central server. The server aggregates these updates and sends the new global model back to the clients who complete the global training round \cite{mcmahan_communication-efficient_2017}.
This learning paradigm has wide applications in industries that face privacy issues when collecting raw data, such as the energy distribution industry~\cite{zhu_review_2022}.

\begin{figure}[t]
    \centerline{\includegraphics[width=\columnwidth]{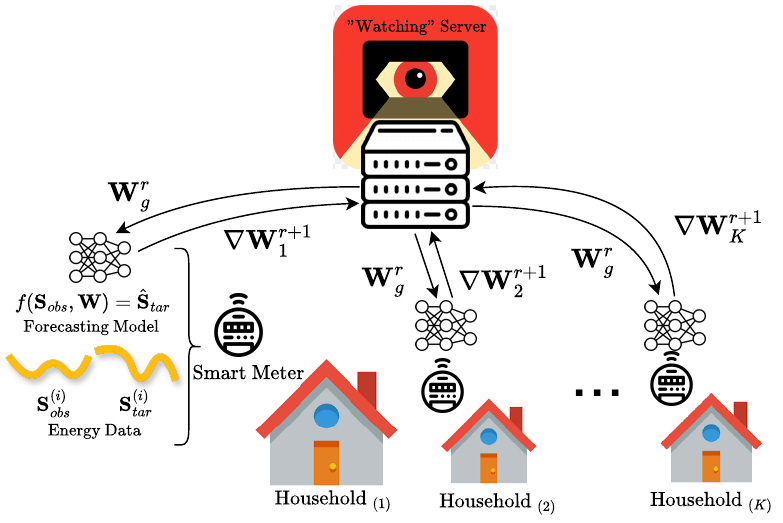}}
    \caption{Federated time series forecasting: a use case in energy demand forecasting with an \textit{honest-but-curious} "watching" server. Each smart meter has a forecasting model and a sequence of energy data. The server distributes the global model parameters and retrieves the locally trained models.}
    \label{fig:fl_senario_ts_inverse}
\end{figure}

The energy industry actively researches data-driven technologies for load forecasting, a typical challenge in time series (TS) data, where historical data is used to predict future load patterns \cite{zhu_review_2022, ahmad_load_2022}.  Specifically the forecasting of individual loads on low-voltage $\text{(smart-)grids}$ is a key subject of interest within the community as it allows the Distribution System Operators to optimize the grid on a local level \cite{ahmad_load_2022, causevic_lv_2023}. 
Local-level forecasting, while beneficial, faces a legal challenge: The EU general data protection regulation (GDPR) protects the privacy of household and corporate energy data, preventing its central aggregation for ML model training \cite{causevic_flexibility_2021, truong_privacy_2021}. This is where FL comes into play, allowing the data to remain at its source. Figure~\ref{fig:fl_senario_ts_inverse}  illustrates an example scenario where households are equipped with smart meters that record private energy data, participating as clients in the FL system.
The objective of the FL system is to build a global forecasting model that predicts individual household energy consumption, using the historical energy consumption data from all clients.

Nonetheless, FL is not without privacy risks, such as membership inference, and property inference attacks \cite{lyu_privacy_2022, rodriguez-barroso_survey_2023}. Given strict privacy regulations like GDPR, it is crucial to ensure that private energy data is not leaked through FL systems \cite{truong_privacy_2021}, just as with sensitive medical images \cite{nguyen_federated_2023}. This paper specifically addresses gradient inversion attacks (GIAs), where attackers aim to reconstruct private training data from client updates~\cite{huang_evaluating_2021} by matching clients' gradients with so-called dummy gradients. Previous research has demonstrated the feasibility of such attacks in image classification tasks, using model parameter updates from FL training to reconstruct training data \cite{zhu_deep_2019, wang_sapag_2020, zhao_idlg_2020, geiping_inverting_2020, deng_tag_2021, yin_see_2021, jeon_gradient_2021, geng_towards_2022, scheliga_dropout_2022, wu_learning_2023}. Improving privacy attacks is essential, as understanding these enables a more rigorous evaluation of defense mechanisms, ultimately enhancing the overall security of privacy preserving systems.

In this paper, we address this gap by focusing on GIAs for federated time series forecasting (TSF) models. Inverting TS poses unique challenges compared to inverting image data with classification tasks. The training data of TS is composed of a sequence of observations and target values, whereas the existing GIA deals with input pairs of images and labels. Reconstructing TS thus requires inverting gradients into the sequences of observations and targets. The network architecture for TSF includes components to capture time dependency, e.g., Gated Recurrent Unit (GRU)~\cite{cho_learning_2014} and 1-D temporal convolution (TCN)~\cite{bai_empirical_2018}. To demystify the inversion risk for TSF, we start with an empirical analysis by applying existing GIAs on combinations of 4 TSF models and 4 datasets. Based on the insights observed, we propose a novel and effective inversion framework, \algo, which utilizes additional knowledge and TS related characteristics for regularization. We also derive the analytical inversion of target values for the special case with a batch size of one. Extensive evaluation results against existing GIAs and ablation studies show that \algo is able to reconstruct both the observation and target values of time series, achieving low reconstructing errors.



Our contributions can be summarized as follows: 
\begin{itemize}
\item We conduct the first empirical analysis of GIAs on federated TSF models, highlighting the challenges of inverting both observation and target sequences with respect to model architecture, gradient distance, and regularization. (Section~\ref{sec:03_empirical_inversion_analysis_on_time_series_forecasting})

\item We design \algo, which includes two innovative components: the gradient inversion model and TS-regularized inversion optimization. The gradient inversion model provides relevant quantile bounds by leveraging auxiliary data. The gradient distance is the L1-Norm with TS regularization incorporating periodicity, trend, and the learned quantile bounds of the gradient inversion model. Additionally, a one-shot analytical technique is used for reconstructing targets. (Section~\ref{sec:04_ts-inverse})

\item We extensively evaluate \algo on four datasets and five forecasting architectures, demonstrating a 2x-10x reduction in reconstruction error on the non-RNN-based architectures. (Section~\ref{sec:05_evaluation})


\end{itemize}


\section{Background}
\subsection{Time Series Forecasting with Federated Learning}
Most of the existing FL studies focus on image and text classification tasks~\cite{rodriguez-barroso_survey_2023}, but its application to TS data is relatively under explored and primarily concentrated in the energy forecasting domain~\cite{fekri_deep_2021, savi_short-term_2021, duttagupta_exploring_2023}. 
In the context of FL for TSF, Recurrent Neural Networks (RNNs)~\cite{rumelhart_learning_1986} and TCNs~\cite{bai_empirical_2018} are commonly used, opposed to traditional statistical models, due to their superior ability to capture complex temporal patterns in data. RNNs~\cite{rumelhart_learning_1986}, specifically Long Short-Term Memory (LSTM) networks~\cite{hochreiter_long_1997} and GRUs~\cite{cho_learning_2014}, are commonly used in federated settings for energy forecasting tasks~\cite{fekri_deep_2021, alhussein_hybrid_2020}. GRUs are an alternative to LSTMs that use fewer gating mechanisms to control flow of information, offering more computational efficiency while maintaining performance.

GRUs are designed to capture dependencies in sequential data by maintaining a hidden state. GRUs consist of two main gates: the update gate and the reset gate. The update gate controls the degree to which the hidden state is updated, while the reset gate determines how much of the previous hidden state should be forgotten. This gating mechanism allows GRUs to capture temporal dependencies efficiently, although it results in a many-to-one mapping from input sequences to hidden states.

Recently, non-recurrent architectures such as TCNs~\cite{bai_empirical_2018} have shown effectiveness in modeling sequence data for TSF tasks by using causal convolutions and dilations to capture long-range temporal dependencies. TCNs offer advantages in parallelization, training stability, and can maintain the same sequence length through zero-padding while using dilations to exponentially increase the receptive field, allowing them to model temporal dependencies more effectively than traditional convolutional neural networks (CNNs) and without the need for recurrence~\cite{lara-benitez_temporal_2020}. Alongside these, simpler Fully Connected Networks (FCNs) are also considered due to their ability to be trained on lower-quality hardware~\cite{duttagupta_exploring_2023}. Despite advancements in transformer-based models for time series analysis, their application in federated TSF remains limited. Therefore, in this paper, we focus on the GRU and TCN architectures for federated TSF.

\subsection{Non-IID Time Series Data in Federated Learning}
Inherited from the nature of FL, training federated TSF models also involves non-identical and independently distributed client data~\cite{fekri_deep_2021, savi_short-term_2021, bousbiat_crossing_2023}. Taking energy load forecasting as an example, each client, such as households and company buildings, possesses a distinct load profile. Each client, which is also the private individual, only contains its own unique load data. It results in the energy load of different clients varying in both peak/off hours and amount. This is different from cases where multiple private individuals can reside in a single client data silo. Although the non-iidness of load profiles and their accompanying convergence issues are beyond the scope of this paper, it is crucial to note that each client's load profile is private and must be protected against privacy leakage~\cite{fernandez_privacy-preserving_2022, ustundag_soykan_differentially_2019}. Notably, the mentioned studies merely discuss differential privacy budgets and accuracy differences, without evaluating any actual attacks~\cite{fernandez_privacy-preserving_2022, ustundag_soykan_differentially_2019}.

\subsection{Privacy and Inversion Attacks on Federated Learning} 
While adversarial attacks in FL aim to modify and disrupt learning tasks~\cite{rodriguez-barroso_survey_2023}, privacy attacks focus on inferring private information from the system~\cite{rodriguez-barroso_survey_2023}. Examples of such privacy attacks include membership inference attacks~\cite{nasr_comprehensive_2019} and property inference attacks~\cite{mo_layer-wise_2021}. Among these privacy attacks, data reconstruction attacks are particularly concerning as they attempt to recover training samples from clients~\cite{zhu_deep_2019, wang_sapag_2020, zhao_idlg_2020, geiping_inverting_2020, deng_tag_2021, yin_see_2021, jeon_gradient_2021, geng_towards_2022, scheliga_dropout_2022, wu_learning_2023}. These attacks, known as GIAs, exploit gradients collected from clients, making them a significant threat to privacy in FL systems. To our knowledge there are no other studies that investigate the GIAs for TS data and models. Therefore, this paper will demonstrate 4 related studies on TS data and models in order to investigate their performance.

The four baselines examined in this study are DLG, InvG, LTI, and DIA. ``Deep Leakage from Gradients'' (DLG)~\cite{zhu_deep_2019} is the pioneering approach that demonstrated how training data could be reconstructed by minimizing the Euclidean distance between gradients, highlighting the feasibility of such attacks. Building upon DLG, ``Inverting Gradient'' (InvG)~\cite{geiping_inverting_2020} introduced the use of cosine similarity and, image processing related, total variation regularization, which improved the reconstruction quality. ``Learning to Invert'' (LTI)~\cite{wu_learning_2023} represents a more recent advancement, leveraging a model trained on an auxiliary dataset to directly map the client samples from gradients, making it robust even against defenses like differential privacy. Lastly, the ``Dropout Inversion Attack`` (DIA)~\cite{scheliga_dropout_2022} extended the attack capabilities to scenarios where dropout layers are used, effectively circumventing this commonly employed defense mechanism. Together, these baselines provide a comprehensive perspective on the effectiveness of GIAs in various federated learning contexts.
\section{Empirical Inversion Analysis on Time Series Forecasting}
\label{sec:03_empirical_inversion_analysis_on_time_series_forecasting}
In this section, we introduce and analyze the risk of inverting clients' private TS data in the context of federated forecasting by applying existing classification-based attacks. We first define TSF and outline the assumptions made regarding the FL setups and threat model. Afterward, we formally define the gradient inversion attacks under our forecasting scenarios. Finally, we analyze the feasibility and risks of existing image-based inversion attacks on different time series models. 

\subsection{Problem Definition}
\subsubsection{Time Series Forecasting} 
\label{sec:problem_definition_time-series_forecasting}

\begin{figure}[t]
    \centering
    \resizebox{\linewidth}{!}{
    \includegraphics[]{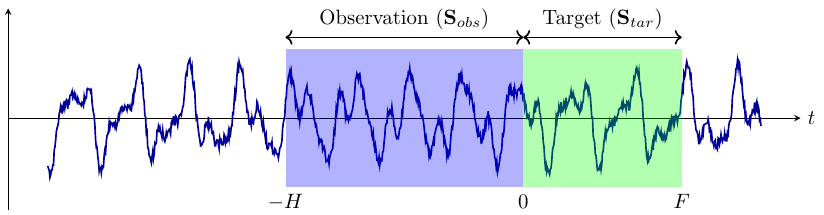}
    }
    \caption{Example of univariate time series forecasting.}
    \label{fig:time-series-univariate}
\end{figure}

TS data are generally represented through discrete time steps, determined by a chosen sampling frequency, though time is inherently continuous. 
TSF is a regression task that predicts future values given a historical segment. In this work, we focus on multi-step TSF where the future segment consists of multiple timesteps. 
Formally, the historical segment, comprising $H$ steps, is defined as a series of subsequent observation timesteps $obs = \left\{t \in \mathbb{Z} \mid -H < t \leq 0\right\}$, whereas the future segment, spanning $F$ steps, is denoted as the targets $tar = \left\{t \in \mathbb{Z} \mid 0 < t \leq F\right\}$ as illustrated in Figure~\ref{fig:time-series-univariate}. We represent historical data of the sequence $\mathbf{S}$ as $\mathbf{S}_{obs} \in \mathbb{R}^{H \times d}$, and the forecasting target data as $\mathbf{S}_{tar} \in \mathbb{R}^{F  \times d}$, where $d$ indicates the number of distinct features. A time series is univariate when $d = 1$, and multivariate when $d > 1$. Furthermore, $\mathbf{S}_t \in \mathbb{R}^d$ is used to denote a value vector at timestep $t$. We focus on the univariate case.
For training the TSF task, we consider recent advanced deep learning models. The training objective of those models is to minimize the mean square errors (MSE) or mean absolute errors (MAE).

\subsubsection{FL for Time Series Forecasting}
\label{sec:problem_definition_federated_learning}
A federated forecasting framework involves $K$ clients, each having its own data silo with sequence data $\mathbf{S}_k$. Each client's sequence $\mathbf{S}_k$ is split up in multiple subsequent pairs of observation data $\mathbf{S}_{obs}^{(i)}$ and target data $\mathbf{S}_{tar}^{(i)}$, where $i$ denotes the $i$'th sample from the sequence.
A client's data silo contains a dataset with sequence samples organized as $\{\mathbf{S}_{obs}^{(i)},\mathbf{S}_{tar}^{(i)}\}_{i=1}^{n_k}$, where $n_k$ denotes the number of local sequence pairs. The clients jointly train a global forecasting model via a strategy determined by the server, e.g., learning rate and batch sizes.

The server initializes a global model with parameters $\mathbf{W}_g^r$, where $g$ stands for the global model and $r$ denotes the global training round. The server distributes $\mathbf{W}_g^r$ to the clients, who then follow federated stochastic gradient decent (FedSGD)~\cite{mcmahan_communication-efficient_2017} to further train the model using their own data. Specifically, the clients perform a single gradient update on a single batch, resulting in gradients $\nabla\mathbf{W}_k^r$. These are sent back to the server, which in turn, aggregates all model updates as $\mathbf{W}_g^{r+1} = \mathbf{W}_g^r - \alpha \left( \frac{1}{K} \sum_{k=1}^K \nabla\mathbf{W}_k^r\right)$, where $\alpha$ is the global learning rate. 

\subsection{Threat model}
In line with the related studies, the threat model for our gradient inversion on TSF models assumes an \textit{honest-but-curious} server. This means that while the server follows protocol rules, it is interested in extracting training data client information from the gradients data it receives.
The attack considered is a \textit{white-box} attack, where the FL strategy, model architecture, optimizer, loss function, observations, and target size ($\mathbb{R}^{d \times H}, \mathbb{R}^{d \times F}$), learning rate ($\alpha$), batch size ($\mathcal{B}$), etc. are all known to the adversary. The adversary has access to the model weights ($\mathbf{W}_{g}^r$) and aggregated gradients ($\nabla \mathbf{W}_k^r$) at communication time and aims to reconstruct each sample from the batched data, including both observations and targets ($\mathbf{S}_{obs}, \mathbf{S}_{tar}$). 
We also assume that the adversary has access to an auxiliary dataset ($\mathcal{D}_{aux}$) with a distribution overlapping with the client's data. Furthermore, the adversary has the computational capabilities to train models outside the FL framework.

\begin{figure*}[b]
    \centering
    \includegraphics[width=\textwidth]{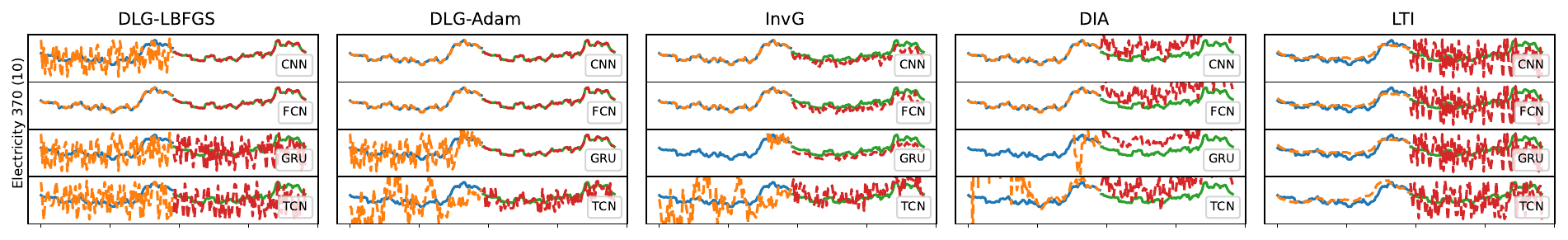}
    \caption{Baseline reconstruction results with FCN, CNN, GRU-2-FCN, and TCN model architectures on the Electricity 370 dataset with seeds (10). In each graph, the blue and green lines are the ground truth observations and targets respectively. The orange and red lines are the respective reconstructed observations and targets.}
    \label{fig:baseline-reconstructions}
\end{figure*}

\subsection{Attacking Time Series Forecasting}
Inverting training data for TSF presents several new challenges compared to inverting training data for classification tasks. The training objectives and inputs differ: for image classification, the objective is cross-entropy loss with pairs of images and their class labels, whereas for time series forecasting, it is mean squared error (MSE) loss with pairs of observation and target series.
In image classification, due to the cross-entropy loss, the labels can be analytically inferred from the gradients, as demonstrated by \textcite{zhao_idlg_2020, geng_towards_2022}. Consequently, the inversion attack on a classification model primarily involves reconstructing the training images.
However, inverting the training data for time series forecasting requires reconstructing both the input observations and the target values ($\mathbf{S}_{obs}$ and $\mathbf{S}_{tar}$) as both are of the same data modality. Furthermore, the analytical method of inferring the labels for the classification task is not directly applicable for a TSF task.
 
To recover the observation and target series data, the adversary first initializes a dummy input ($\widetilde{\mathbf{S}}_{obs}$) and target ($\widetilde{\mathbf{S}}_{tar}$) pair with the same dimensions as $\mathbf{S}_{obs}$ and $\mathbf{S}_{tar}$, respectively, and calculates the corresponding gradients:
\begin{equation}
    \nabla \widetilde{\mathbf{W}} = \frac{\partial \mathcal{L} (f(\widetilde{\mathbf{S}}_{obs}, \mathbf{W}), \widetilde{\mathbf{S}}_{tar})}{\partial \mathbf{W}}
    \label{eq:dummy_gradients},
\end{equation}
where $f$ is the global model and $\mathbf{W}$ are its parameters before applying the gradient updates. $f$ is a twice differentiable neural network and $\mathcal{L}$ is the regression loss function used in optimizing $f$. These dummy observations and targets are optimized by minimizing a distance $D$ between the dummy gradients $\nabla \widetilde{\mathbf{W}}$ from \eqref{eq:dummy_gradients} and clients' gradients $\nabla \mathbf{W}$:
\begin{equation}
    (\widetilde{\mathbf{S}}_{obs}^{\star}, \widetilde{\mathbf{S}}_{tar}^{\star}) = \underset{ \widetilde{\mathbf{S}}_{obs}, \widetilde{\mathbf{S}}_{tar}}{\arg\min} \, D(\nabla \widetilde{\mathbf{W}}, \nabla \mathbf{W}).
\end{equation}

In this work, we define the optimization problem for recovering both the observations ($\mathbf{S}_{obs}$) and the targets ($\mathbf{S}_{tar}$). Specifically, we use the gradient to recover data for both observations and targets. Additionally, the TS data is often pre-processed to be within the domain [0, 1], which helps to bound the reconstructed data.
DLG \cite{zhu_deep_2019} and others \cite{zhao_idlg_2020, deng_tag_2021, geng_towards_2022, yin_see_2021} uses L2-Norm to match the gradients as:
\begin{equation}
    D(\nabla \widetilde{\mathbf{W}}, \nabla \mathbf{W}) = \left\| \nabla \widetilde{\mathbf{W}} - \nabla \mathbf{W} \right\|_2.
\end{equation}
Inverting Gradients \cite{geiping_inverting_2020} and others \cite{scheliga_dropout_2022, jeon_gradient_2021} use the Cosine Similarity loss together with the image prior regularization, total variation, $\mathcal{R}_{\text{TV}}$, with hyperparameter $\lambda_{\text{TV}}$. The Cosine Similarity loss is defined as
\begin{equation}
    \mathcal{D}_{\text{cosine}}(\nabla \widetilde{\mathbf{W}}, \nabla \mathbf{W}) = 1 - \frac{\nabla \widetilde{\mathbf{W}} \cdot \nabla \mathbf{W}}{\|\nabla \widetilde{\mathbf{W}}\|_2 \|\nabla \mathbf{W}\|_2} + \lambda_{\text{TV}} \mathcal{R}_{\text{TV}}.
\end{equation}

\subsection{Empirical Findings}
We present our empirical findings by applying four existing attacks---DLG~\cite{zhu_deep_2019}, InvG~\cite{geiping_inverting_2020}, DIA~\cite{scheliga_dropout_2022}, and LTI~\cite{wu_learning_2023}--- on three models: CNN, FCN, and TCN for inverting the TS data from FL gradients. These attacks differ in their methodologies and gradient distance functions. Specifically, we discuss the impact of model architecture, loss function, and total variation for inversion performance. The experimental setups are detailed in Section~\ref{sec:experiment_setup}. We use the Symmetrical Mean Absolute Percentage Error (sMAPE$\downarrow$) as the evaluation metric to compare all baselines, presented in Table \ref{tab:big_comparison}. 
Specifically, to demonstrate the impact of the total variation regularization ($\mathcal{R}_{\text{TV}}$) for InvG, we report the sMAPE results in Table \ref{tab:invg_reconstructions_with_total_variation}.
Additionally, we also visualize the recovered observation and target data for easy perceptual comparison, by a single TS example in Figure \ref{fig:baseline-reconstructions}. In-depth comparison is summarized in Table~\ref{tab:big_comparison} and analyzed in Section~\ref{sec:05_evaluation}.

\subsubsection{Model Architectures}
The TCN architecture \cite{bai_empirical_2018} includes dropout layers, which make the optimization attack more difficult \cite{zheng_dropout_2022}.  DIA~\cite{scheliga_dropout_2022} counters this dropout defense by also optimizing the dropout masks that indicate which neurons are dropped during the training round. This is required in order to accurately reconstruct training data from architectures with dropout layers. 
Moreover, GRUs are designed to selectively remember and forget information, a process controlled through nonlinear functions governed by reset and update gates \cite{cho_learning_2014}. These gates determine the flow and modification of information over time, inherently leading to a loss of information and many-to-one mappings from input sequences to hidden states.
From a gradient perspective, while gradients provide insights into how changes in inputs could affect the outputs, they only do so on average over the whole input space and do not offer direct mappings to individual inputs. This many-to-one relationship also applies to RNN architectures.

\subsubsection{Gradient Distance Function of Attacks}
The effects of the gradient distance function are most clear for the CNN and FCN architectures, according to Table~\ref{tab:big_comparison} and Figure~\ref{fig:baseline-reconstructions}. 
The DLG attack, which uses L2-Norm to align gradients, effectively reconstructs the observations and targets. This is because L2-Norm measures the squared differences between gradients, inherently capturing both the magnitude and the direction of the gradients. In contrast, the InvG and DIA attacks rely on Cosine Similarity, which normalizes the gradients and focuses only on the direction, disregarding their magnitude. This focus on direction alone results in noisier targets, indicating that the magnitude of gradients is essential for accurate target reconstruction.

\subsubsection{Total Variation}
The application of total variation \cite{geiping_inverting_2020}, is observed to have a counterproductive effect on the reconstruction of TS data, as shown in Table \ref{tab:invg_reconstructions_with_total_variation}. The regularization results in a higher sMAPE value when applied to TS. We applied the total variation separately to the observations and targets using hyperparameters $\lambda_{\text{TV}}^{obs}$ and $\lambda_{\text{TV}}^{tar}$, respectively. When increasing $\lambda_{\text{TV}}^{obs}$ while keeping $\lambda_{\text{TV}}^{tar}=0$, the sMAPE increased for the observations, while for the targets it remained the same. In contrast, when increasing $\lambda_{\text{TV}}^{tar}$ while keeping $\lambda_{\text{TV}}^{obs}=0$, the sMAPE for observations increased, while it decreased for targets. The total variation reduces noise in images, which helps improve the image reconstruction quality. However, this does not translate well to TS data. This is because TS data can be inherently noisier than images due to large differences between subsequent timesteps.

\highlight{
\begin{itemize}
\item TCN and GRU architectures are more challenging to invert compared to CNN and FCN due to their inherent design features like dropout layers and gate mechanisms.
\item  To correctly recreate targets, the gradient distance has to take into account the magnitudes of the gradient values. 
\item Image regularization priors are not effective for time series data, which tends to be intrinsically noisier and exhibit larger differences between subsequent timesteps.
\end{itemize}
}

\begin{table}[]
\caption{Comparison of the effects of total variation regularization on the InvG attack, measured using sMAPE ($\downarrow$) with a CNN architecture. Evaluations are performed on the Electricity 370 and Proprietary Dataset for different combinations of $\lambda_{\text{TV}}^{obs}$ and $\lambda_{\text{TV}}^{tar}$ values. The experiments are run with a batch size $\mathcal{B} = 1$ and seeds 10, 43, and 28, with the best performances highlighted in bold. Standard deviations are provided as subscripts.}
\label{tab:invg_reconstructions_with_total_variation}

\centering
\resizebox{\linewidth}{!}{
\begin{tabular}{l|l|cc|cc}
\toprule
      & Dataset                 & \multicolumn{2}{c}{Electricity 370} & \multicolumn{2}{c}{Proprietary Dataset} \\
\midrule
$\lambda_{\text{TV}}^{obs}$ & $\lambda_{\text{TV}}^{tar}$ & Observation & Target & Observation & Target \\
\midrule
0	& 0	 & \textbf{0.002$_{0.00}$} & 0.094$_{0.06}$ & \textbf{0.016$_{0.00}$} & 1.143$_{0.23}$ \\
	& 0.001	 & 0.100$_{0.01}$ & 0.042$_{0.04}$ & 0.733$_{0.16}$ & 1.040$_{0.25}$ \\
	& 0.01	 & 0.797$_{0.04}$ & \textbf{0.009$_{0.00}$} & 1.528$_{0.07}$ & 0.036$_{0.02}$ \\
\midrule
0.001	& 0	 & 0.029$_{0.01}$ & 0.094$_{0.06}$ & 0.170$_{0.02}$ & 1.144$_{0.23}$ \\
	& 0.001	 & 0.067$_{0.01}$ & 0.042$_{0.04}$ & 0.467$_{0.14}$ & 1.042$_{0.25}$ \\
	& 0.01	 & 0.501$_{0.06}$ & \textbf{0.009$_{0.00}$} & 1.394$_{0.06}$ & \textbf{0.028$_{0.01}$} \\
\midrule
0.01	& 0	 & 0.029$_{0.00}$ & 0.094$_{0.06}$ & 0.165$_{0.02}$ & 1.145$_{0.23}$ \\
	& 0.001	 & 0.067$_{0.00}$ & 0.042$_{0.04}$ & 0.472$_{0.13}$ & 1.042$_{0.25}$ \\
	& 0.01	 & 0.500$_{0.06}$ & \textbf{0.009$_{0.00}$} & 1.398$_{0.06}$ & \textbf{0.028$_{0.01}$} \\
\bottomrule
\end{tabular}}

\end{table}

\section{\algo Framework}
\label{sec:04_ts-inverse}
\begin{figure*}[t]
    \centering
    \includegraphics[width=\textwidth]{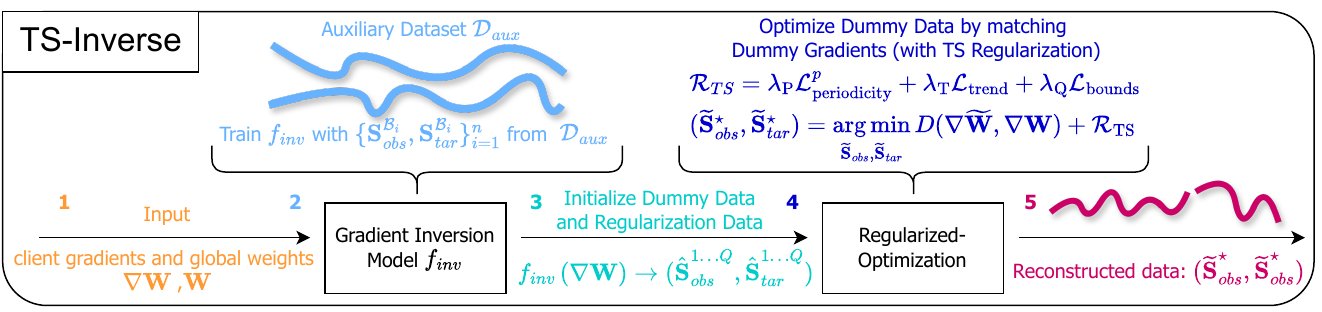}
    \caption{Key steps of \algo framework}
    \label{fig:ts-inverse-framework}
\end{figure*}

Following the findings from the empirical study, we propose a first-of-kind gradient inversion attack for federated TSF, named \algo, which inverts a batch of observation and target data of time series regression tasks from clients gradients. We illustrate the framework of \algo in Figure~\ref{fig:ts-inverse-framework}, consisting of three key steps. The first component of \algo is to train a gradient inversion model, i.e., $f_{inv}$, using auxiliary data to map model gradients to sequences of values that correspond to specific quantile ranges of the observations and targets, as depicted in Figure~\ref{fig:ts-inverse-quantile-reconstructions}. The quantile ranges divide the data into intervals with equal probabilities, giving a representation of the observation and target distribution that belongs to the gradients. These sequences serve as upper and lower bounds for the learned quantile regularization. The second key feature is the inversion optimization, starting from the dummy time series. Different from the convention of minimizing L2-Norm and cosine distance in image inversion attacks, we advocate using L1-Norm loss and combine it with the proposed time series specific regularization terms, namely periodicity, trend, and quantile bounds. Furthermore, we provide an analytical one-shot inversion technique for the target sequence, under the scenario where the batch size is 1. 

\subsection{Gradient Inversion Model}
\label{sec:04_01_grad_inversion_model}
We assume the existence of an auxiliary dataset $\mathcal{D}_{aux}$ by the \textit{honest-but-curious} server. This is ideally a subset of the time series of the data silos that are being attacked, but it can also be an out-of-distribution dataset \cite{wu_learning_2023}.
The dataset is first pre-processed such that the samples have the same sequence length as the client's data samples through interpolation. 

We define a gradient inversion model $f_{inv}: \mathbb{R}^m \rightarrow (\mathbb{R}^{H \times d \times Q}, \mathbb{R}^{F \times d \times Q})$, where $m$ is the number of parameters in the forecasting model $f$, and $Q$ is the number of quantile bounds. This model is designed to map gradients to quantile bounds, which are sequences of values corresponding to specific quantile levels of the TS observations and targets. Quantile levels ($\tau$) are probability values between 0 and 1. For example, between the bounds corresponding to quantile levels $0.1$ and $0.9$ fall $80\%$ of the sequences.

The model $f_{inv}$ is trained on the gradients of the forecasting model $f$ derived from batches of samples from $\mathcal{D}_{aux}$. A batch $(\mathbf{S}_{obs}^{\mathcal{B}}, \mathbf{S}_{tar}^{\mathcal{B}})$ consists of observation and target pairs from $\mathcal{D}_{aux}$, where $\mathcal{B}$ denotes the number of such pairs.

The model $f_{inv}$ is optimized using the pinball loss with respect to the batched samples. It produces the outputs $\{\hat{\mathbf{S}}_{obs}^{\tau_1}, \hat{\mathbf{S}}_{obs}^{\tau_2}, \dots, \hat{\mathbf{S}}_{obs}^{\tau_Q}\}$ and $\{\hat{\mathbf{S}}_{tar}^{\tau_1}, \hat{\mathbf{S}}_{tar}^{\tau_2}, \dots, \hat{\mathbf{S}}_{tar}^{\tau_Q}\}$, where each $\tau_q$ represents a specific quantile level. These quantile predictions are repeated along the batch dimension to match $(\mathbf{S}_{obs}^{\mathcal{B}}, \mathbf{S}_{tar}^{\mathcal{B}})$. As a result the quantile predictions demonstrate the variability of the batch. 
The pinball loss, which is used for optimization, is defined as follows:
\begin{equation}
\mathcal{L}_{\text{pinball}}^{\tau}(\mathbf{S}, \hat{\mathbf{S}}) = \frac{1}{T} \sum_{t=1}^{T} \max\left((\tau - 1)(\mathbf{S}_t - \hat{\mathbf{S}}_t), \tau (\mathbf{S}_t - \hat{\mathbf{S}}_t) \right),
\label{eq:pinball}
\end{equation}
where $t$ is a timestep and $T$ is the total number of timesteps in sequence $\mathbf{S}$, and $\hat{\mathbf{S}}$ represents the predicted quantile bound. The total loss is to apply  \eqref{eq:pinball} for each quantile level over the observations and targets as:
\begin{equation}
    \mathcal{L}_{inv} = \sum_{q=1}^Q \frac{\mathcal{L}_{\text{pinball}}^{\tau_q}(\mathbf{S}_{obs}^{\mathcal{B}}, \hat{\mathbf{S}}_{obs}^{\tau_q}) + \mathcal{L}_{\text{pinball}}^{\tau_q}(\mathbf{S}_{tar}^{\mathcal{B}}, \hat{\mathbf{S}}_{tar}^{\tau_q})}{2}.
\end{equation}

By using the pinball loss function, the model $f_{inv}$ is effectively trained to provide quantile bounds for the entire batch of TS data.

\subsection{Gradient Matching Distance Function}
\label{sec:04_02_gradient_distance}
The cosine similarity loss only matches the hyperplane direction of the gradients, but does not take into account the magnitudes of the values. 
Matching the magnitude of the values is essential for gradient approximation. Otherwise, the optimization will be dominated by the large values, resulting in distortion of updating dummy data.
One can add an additional term to the cosine similarity loss that takes into account the norm of the difference of the gradients \cite{jiang_delving_2022, zhang_two_2024}. 

On the other hand, the L1 loss, i.e., the summed absolute errors, is a single-term metric that is robust to outliers, which is common in gradients. Additionally. it matches the magnitude of the gradient values and it is particularly compatible with models that utilize weights initialized from a normal distribution \cite{deng_tag_2021}, such as the TCN architecture. Our choice is backed by empirical findings that show that using the L1-Norm leads to better reconstructions compared to the Cosine with or without the L2-Norm. The distance function is thus formulated as follows:
\begin{equation}
    D(\nabla \widetilde{\mathbf{W}}, \nabla \mathbf{W}) = \left\|\nabla \widetilde{\mathbf{W}} - \nabla \mathbf{W}\right\|_1.
    \label{eq:gradient_distance}
\end{equation}

\begin{table*}[]
\caption{Comparison of sMAPE (↓) for baseline attacks and \algo on multiple datasets (Electricity 370, KDDCup, London Smartmeter, and Proprietary Dataset) and model architectures (CNN, FCN, TCN). Each attack is conducted over 5000 steps with a batch size \(\mathcal{B} = 1\). The attack methods include 'DLG-LBFGS', 'DLG-Adam', 'InvG', 'DIA', 'LTI', '\algo', and '\algo\(_{\text{one-shot}}\)'. Experiments are run with seeds 10, 43, 28, 80, and 71, with the best performances highlighted in bold. Standard deviations are provided as subscripts.}
\label{tab:big_comparison}
\centering
\resizebox{\linewidth}{!}{
\begin{tabular}{l|l|cc|cc|cc|cc}
\toprule
      & Dataset                 & \multicolumn{2}{c}{Electricity 370} & \multicolumn{2}{c}{KDDCup} & \multicolumn{2}{c}{London Smartmeter} & \multicolumn{2}{c}{Proprietary Dataset} \\
\midrule
Model & Attack Method & Observation & Target           & Observation & Target           & Observation & Target           & Observation & Target \\
\midrule
CNN	& DLG-LBFGS	 & 0.632$_{0.05}$ & 0.001$_{0.00}$ & 1.454$_{0.11}$ & 0.030$_{0.03}$ & 1.429$_{0.14}$ & 0.023$_{0.01}$ & 1.294$_{0.08}$ & 0.011$_{0.00}$ \\
	& DLG-Adam	 & 0.029$_{0.03}$ & 8.6e-05$_{0.00}$ & 1.349$_{0.23}$ & 0.029$_{0.03}$ & 0.993$_{0.14}$ & 0.006$_{0.00}$ & 1.013$_{0.27}$ & 0.005$_{0.00}$ \\
	& InvG	 & 0.002$_{0.00}$ & 0.189$_{0.13}$ & 0.038$_{0.02}$ & 1.306$_{0.30}$ & 0.027$_{0.00}$ & 1.136$_{0.26}$ & 0.015$_{0.00}$ & 1.160$_{0.19}$ \\
	& DIA	 & 0.002$_{0.00}$ & 0.542$_{0.11}$ & 0.073$_{0.04}$ & 1.293$_{0.11}$ & 0.094$_{0.04}$ & 1.501$_{0.05}$ & 0.018$_{0.01}$ & 1.484$_{0.08}$ \\
	& LTI	 & 0.140$_{0.04}$ & 0.685$_{0.07}$ & 0.889$_{0.15}$ & 1.399$_{0.28}$ & 0.713$_{0.18}$ & 1.469$_{0.06}$ & 0.486$_{0.05}$ & 1.305$_{0.09}$ \\
	& \algo	 & \textbf{0.002$_{0.00}$} & 1.2e-05$_{0.00}$ & \textbf{0.017$_{0.01}$} & 2.1e-04$_{0.00}$ & \textbf{8.1e-05$_{0.00}$} & 2.4e-05$_{0.00}$ & \textbf{0.009$_{0.00}$} & 7.1e-05$_{0.00}$ \\
 & \algo$_{\text{one-shot}}$	 & 0.003$_{0.00}$ & \textbf{1.4e-07$_{0.00}$} & 0.085$_{0.05}$ & \textbf{1.9e-06$_{0.00}$} & 0.024$_{0.01}$ & \textbf{2.1e-06$_{0.00}$} & 0.032$_{0.01}$ & \textbf{7.3e-07$_{0.00}$} \\
\midrule
FCN	& DLG-LBFGS	 & 0.025$_{0.01}$ & 2.0e-04$_{0.00}$ & 0.168$_{0.05}$ & 0.002$_{0.00}$ & 0.498$_{0.17}$ & 0.003$_{0.00}$ & 0.181$_{0.07}$ & 0.002$_{0.00}$ \\
	& DLG-Adam	 & 2.8e-06$_{0.00}$ & 1.2e-05$_{0.00}$ & 0.002$_{0.00}$ & 4.1e-04$_{0.00}$ & 0.004$_{0.00}$ & 2.9e-05$_{0.00}$ & 1.8e-05$_{0.00}$ & 3.5e-05$_{0.00}$ \\
	& InvG	 & 4.7e-06$_{0.00}$ & 0.214$_{0.15}$ & 6.1e-05$_{0.00}$ & 1.261$_{0.39}$ & 7.8e-05$_{0.00}$ & 1.300$_{0.10}$ & 2.6e-05$_{0.00}$ & 1.108$_{0.15}$ \\
	& DIA	 & 0.004$_{0.00}$ & 0.542$_{0.10}$ & 0.027$_{0.02}$ & 1.343$_{0.19}$ & 0.068$_{0.05}$ & 1.571$_{0.07}$ & 1.1e-04$_{0.00}$ & 1.551$_{0.07}$ \\
	& LTI	 & 0.133$_{0.03}$ & 0.685$_{0.07}$ & 0.604$_{0.17}$ & 1.399$_{0.28}$ & 0.673$_{0.07}$ & 1.469$_{0.06}$ & 0.498$_{0.03}$ & 1.305$_{0.09}$ \\
	& \algo	 & \textbf{1.7e-06$_{0.00}$} & 8.4e-07$_{0.00}$ & \textbf{2.4e-06$_{0.00}$} & \textbf{1.1e-06$_{0.00}$} & \textbf{6.3e-06$_{0.00}$} & 2.5e-06$_{0.00}$ & \textbf{6.4e-06$_{0.00}$} & 4.4e-06$_{0.00}$ \\
	& \algo$_{\text{one-shot}}$	 & \textbf{3.1e-07$_{0.00}$} & \textbf{8.3e-08$_{0.00}$} & 5.3e-06$_{0.00}$ & 1.5e-06$_{0.00}$ & 3.2e-05$_{0.00}$ & \textbf{1.4e-06$_{0.00}$} & 6.6e-06$_{0.00}$ & \textbf{8.1e-07$_{0.00}$} \\
\midrule
TCN	& DLG-LBFGS	 & 0.652$_{0.06}$ & 0.689$_{0.04}$ & 1.483$_{0.06}$ & 1.374$_{0.28}$ & 1.437$_{0.10}$ & 1.429$_{0.09}$ & 1.321$_{0.13}$ & 1.243$_{0.09}$ \\
	& DLG-Adam	 & 0.726$_{0.07}$ & 0.362$_{0.11}$ & 1.320$_{0.13}$ & 1.072$_{0.41}$ & 1.136$_{0.18}$ & 1.079$_{0.19}$ & 1.141$_{0.16}$ & 1.028$_{0.09}$ \\
	& InvG	 & 0.704$_{0.05}$ & 0.553$_{0.20}$ & 1.359$_{0.16}$ & 1.445$_{0.19}$ & 1.154$_{0.18}$ & 1.493$_{0.11}$ & 1.114$_{0.12}$ & 1.424$_{0.11}$ \\
	& DIA	 & 1.088$_{0.31}$ & 0.719$_{0.20}$ & 0.728$_{0.20}$ & 1.229$_{0.07}$ & 1.076$_{0.57}$ & 1.576$_{0.01}$ & 1.282$_{0.34}$ & 1.473$_{0.05}$ \\
	& LTI	 & 0.158$_{0.05}$ & 0.628$_{0.06}$ & 0.839$_{0.13}$ & 1.437$_{0.30}$ & 0.743$_{0.08}$ & 1.402$_{0.06}$ & 0.530$_{0.13}$ & 1.326$_{0.09}$ \\
	& \algo	 & 0.097$_{0.04}$ & 0.007$_{0.01}$ & \textbf{0.509$_{0.13}$} & 0.042$_{0.05}$ & 0.194$_{0.22}$ & 0.106$_{0.19}$ & \textbf{0.172$_{0.05}$} & 0.021$_{0.02}$ \\
	& \algo$_{\text{one-shot}}$	 & \textbf{0.089$_{0.03}$} & \textbf{1.2e-07$_{0.00}$} & 0.559$_{0.21}$ & \textbf{1.3e-06$_{0.00}$} & \textbf{0.188$_{0.12}$} & \textbf{1.8e-06$_{0.00}$} & 0.251$_{0.14}$ & \textbf{7.6e-07$_{0.00}$} \\
\bottomrule
\end{tabular}}

\end{table*}

\subsection{Regularizations: Periodicity, Trend and Quantile Bounds}
\label{sec:04_03_ts_regularization}
In time series, there are data-specific characteristics that can be leveraged to regularize the reconstruction: periodicity and trend.  The former refers to the repeating patterns or cycles in the data over a specific period, whereas the trend represents the long-term progression or direction in the data. Furthermore, the learned quantile ranges, outputted by $f_{inv}$, are used as soft bounds for the reconstructed data.
We weight and combine those patterns  into one regularization term, $\mathcal{R}_{TS}$:
\begin{align}
    \mathcal{R}_{TS} &= \lambda_{\text{P}} \mathcal{L}_{\text{periodicity}}^p + \lambda_{\text{T}} \mathcal{L}_{\text{trend}} + \lambda_{\text{Q}}^{obs}\mathcal{L}_{\text{bounds}}^{obs}  + \lambda_{\text{Q}}^{tar}\mathcal{L}_{\text{bounds}}^{tar},
    \label{eq:ts_regularization}
\end{align}
where $\lambda_{\text{P}}$, $\lambda_{\text{T}}$, $\lambda_{\text{Q}}^{obs}$, and $\lambda_{\text{Q}}^{tar}$ are hyperparameters controlling the importance of periodicity, trend and learned quantile bounds regularizations, respectively.

\subsubsection{Periodicity Regularization} 
Understanding the specific periodic nature of the data is crucial and the periodicity must be present in the combined observation and target sequence.
To capture periodicity, we minimize the absolute error between each point and its corresponding point one period $p$ apart. Here, $p$ represents the number of timesteps for one full period or cycle. This regularization ensures that reconstructed data maintains the inherent cyclical patterns. Periodicity can vary widely depending on the dataset, exhibiting daily, weekly, or even yearly patterns. The regularization is defined as the mean absolute error between the periodic points:

\begin{equation}
    \mathcal{L}_{\text{periodicity}}^p(\mathbf{S}) = \frac{1}{T-p} \sum_{t=1}^{T-p} \left| \mathbf{S}_t - \mathbf{S}_{t+p} \right|,
    \label{eq:periodicity_regularization}
\end{equation}

where $p$ denotes the period length and $T$ is the total number of timesteps in the sequence $\mathbf{S}$.

\subsubsection{Trend Regularization}
To enforce trend consistency, we fit a linear trend to the sequence and minimize the deviation of the actual sequence from this trend. This approach ensures that the trend is regularized over both the observations and targets combined, capturing the underlying trend dynamics in the data. The regularization is defined as the MAE between the sequence and the trend of the sequence, as such:
\begin{equation}
    \mathcal{L}_{\text{trend}}(\mathbf{S}) = \left\| \mathbf{S} - \mathbf{S}_\text{trend} \right\|_1,
    \label{eq:trend_regularization}
\end{equation}
where the $\mathbf{S}_\text{trend}$ is calculated using linear regression over $\mathbf{S}$ as such: $\mathbf{S}_\text{trend} = \beta \left(s - \bar{s} \right) + \bar{\mathbf{S}}$. Here, $s$ represents the time indices of the sequence, $\bar{s}$ is the mean of these time indices, $\bar{\mathbf{S}}$ is the mean of the sequence $\mathbf{S}$, and the slope $\beta$ of the linear trend is given by: $\beta = \frac{\sum \left( s - \bar{s} \right) \cdot \left( \mathbf{S} - \bar{\mathbf{S}} \right)}{\sum \left( s - \bar{s} \right)^2}$.

\begin{figure*}[t]
    \centering
    \includegraphics[width=\textwidth]{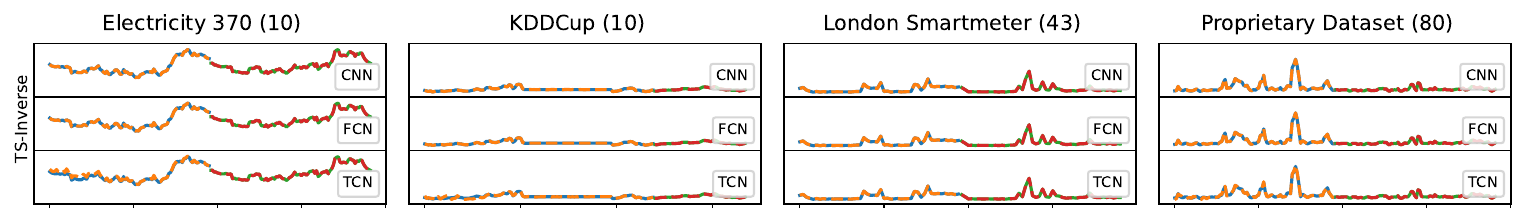}
    \caption{\algo reconstruction results with FCN, CNN, and TCN model architectures on a sample from each dataset.}
    \label{fig:ts-inverse_reconstructions}
\vspace{-20pt}
\end{figure*}

\subsubsection{Quantile Bounds Regularization}
The predictions made by the gradient inversion model are designed to function as soft bounds for the dummy data as illustrated in Figure~\ref{fig:ts-inverse-quantile-reconstructions}. The bounds are defined as quantile pairs that are symmetrically spaced around the median. Within these bounds, no regularization is applied. However, if the dummy data falls outside this range, an absolute error is calculated for regularization. This approach intuitively ensures that the dummy data is encouraged to stay within realistic and statistically consistent bounds.

For a sequence $\mathbf{S}$ and bounds $\mathbf{S}_{\text{lower}}$ and $\mathbf{S}_{\text{upper}}$, the regularization loss is defined as:
\begin{equation}
    \mathcal{L}_{\text{bounds}} = \sum_{q=1}^Q \left\| \max(0, \mathbf{S} - \mathbf{S}_\text{upper}) + \max(0, \mathbf{S}_\text{lower} - \mathbf{S}) \right\|_1,
    \label{eq:bounds_regularization}
\end{equation}
where $q$ indexes over the quantiles with $\mathbf{S}_\text{lower} = \mathbf{S}^{\tau_q}$ and $\mathbf{S}_\text{upper} = \mathbf{S}^{\tau_{(Q-q)}}$, respectively.

Putting Eq. \eqref{eq:periodicity_regularization} \eqref{eq:trend_regularization}, and \eqref{eq:bounds_regularization} into \eqref{eq:ts_regularization} and combining it with \eqref{eq:gradient_distance}, \algo employs the following loss function to invert a batch of time series of observation and target from the client's gradients:
\begin{equation*}
    \mathcal{L}_{\text{total}} = D(\nabla \widetilde{\mathbf{W}}, \nabla \mathbf{W}) + \mathcal{R}_{TS}.
\end{equation*}

\subsection{One-shot Target Reconstruction for $\mathcal{B} = 1$}
\label{sec:04_04_analytical_target_reconstruction}
In this section, we demonstrate how to reconstruct the target predictions of a fully connected layer $f_{FC}$, which in practice is the last layer of the forecasting model $f$.
Given a fully connected layer $f_{FC}$ with weights $\mathbf{W}$ and biases $\mathbf{b}$, and knowing the gradients $\nabla \mathbf{W}$ and $\nabla \mathbf{b}$ obtained with training inputs $x$ and outputs $y$, we assume the objective function is the MSE $\mathcal{L}_{\text{MSE}} = \|\hat{y} - y\|_2$, where $\hat{y}$ are the layer's predictions. It is possible to reconstruct $y$ one-shot for $\mathcal{B} = 1$ if $\nabla \mathbf{b}$ is invertible.
The MSE loss is defined as: $\mathcal{L}_{\text{MSE}} = \frac{1}{N} \sum_{i=1}^{N} (\hat{y}_i - y_i)^2$, where $N$ is the number of elements. We can express the predicted output as $\hat{y} = \mathbf{W}x + \mathbf{b}$.
The gradient with respect to the bias $\mathbf{b}$ and weights $\mathbf{W}$ are:
\begin{align}
    \nabla \mathbf{b} &= \frac{2}{N} (\mathbf{W}x + \mathbf{b} - y) \label{eq:gradient_of_b} \\
    \nabla \mathbf{W} &= \frac{2}{N} (\mathbf{W}x + \mathbf{b} - y)x^T.
    \label{eq:gradient_of_w}
\end{align}
Equation \eqref{eq:gradient_of_b} can be rearranged to solve for $y$:
\begin{equation}
    y = \mathbf{W}x + \mathbf{b} - \nabla \mathbf{b} \cdot \frac{N}{2}.
    \label{eq:y_from_nabla_b}
\end{equation}
Here, $x$ is unknown, but it can be reconstructed by substituting $y$ from \eqref{eq:y_from_nabla_b} into \eqref{eq:gradient_of_w}, we get:
\begin{align*}
    \nabla \mathbf{W} &= \frac{2}{N} \left( \mathbf{W}x + \mathbf{b} - \left( \mathbf{W}x + \mathbf{b} - \nabla \mathbf{b} \cdot \frac{N}{2} \right) \right)x^T \\
    &= \nabla \mathbf{b} \cdot x^T.
\end{align*}
If $\nabla \mathbf{b}$ is invertible, such that $(\nabla \mathbf{b})^{-1}$ exists, then x$^T = (\nabla \mathbf{b})^{-1} \nabla \mathbf{W}$. Finally, substituting $x$ back into equation (\ref{eq:y_from_nabla_b}), we obtain:
\begin{equation}
    y = \mathbf{W}(\nabla \mathbf{b})^{-1} \nabla \mathbf{W} + \mathbf{b} - \nabla \mathbf{b} \cdot \frac{N}{2}.
\end{equation}
This demonstrates that the target $y$ can be reconstructed in one shot for $\mathcal{B} = 1$, provided that $\nabla \mathbf{b}$ is invertible.
\begin{table}[] 
\caption{Dataset descriptions: sample frequency (FREQ.), observation length ($O$), future length ($F$), window size ($W$), and step size for attacked samples ($S_{attack}$) and for the auxilary samples ($S_{aux}$).}
\label{tab:datasets}
\centering
\begin{tabular}{lcccccc} 
\toprule
Dataset           & FREQ.  & $H$ & $F$ & $W$ & $S_{attack}$ & $S_{aux}$ \\
\midrule
Electricity 370   & 15 MIN & 96           & 96  & 192 & 96 & 4    \\
London SmartMeter & 30 MIN & 48           & 48  & 96  & 48 & 2    \\
Proprietary   & 15 MIN & 96           & 96  & 192 & 96 & 4    \\ 
KDDCup 2018       & HOUR   & 120          & 48  & 168 & 24 & 1    \\
\bottomrule
\end{tabular}
\end{table}

\begin{table}[]
\caption{Results of \algo without regularization for the TCN model architecture, measured using sMAPE (↓). Gradient distance functions include 'Cosine + L1', 'Cosine + L2', 'Cosine', 'L2', and 'L1'. Evaluations are performed on the Electricity 370 and London Smartmeter datasets for \(\mathcal{B} \in [1, 2, 4]\). The experiments are run with seeds 10, 43, and 28, with the best performances highlighted in bold. Standard deviations are provided as subscripts.}
\label{tab:ts-inverse_tcn_gradient_loss_comparison} 
\centering
\resizebox{\linewidth}{!}{
\begin{tabular}{l|l|cc|cc}
\toprule
      & Dataset                 & \multicolumn{2}{c}{Electricity 370} & \multicolumn{2}{c}{London Smartmeter} \\
\midrule
$\mathcal{B}$ & Grad. Distance $D$ & Observation & Target           & Observation & Target \\
\midrule
1	& Cosine + L1	 & 0.083$_{0.01}$ & \textbf{9.3e-08$_{0.00}$} & 0.212$_{0.03}$ & \textbf{2.3e-06$_{0.00}$} \\
	& Cosine + L2	 & 0.290$_{0.17}$ & \textbf{9.3e-08$_{0.00}$} & 0.481$_{0.30}$ & \textbf{2.3e-06$_{0.00}$} \\
	& Cosine	 & 0.377$_{0.19}$ & \textbf{9.3e-08$_{0.00}$} & 0.860$_{0.56}$ & \textbf{2.3e-06$_{0.00}$} \\
	& L2	 & 0.236$_{0.05}$ & \textbf{9.3e-08$_{0.00}$} & 0.783$_{0.49}$ & \textbf{2.3e-06$_{0.00}$} \\
	& L1	 & \textbf{0.081$_{0.03}$} & \textbf{9.3e-08$_{0.00}$} & \textbf{0.208$_{0.14}$} & \textbf{2.3e-06$_{0.00}$} \\
\midrule
2	& Cosine + L1	 & \textbf{0.362$_{0.14}$} & \textbf{0.053$_{0.04}$} & 0.301$_{0.20}$ & 0.261$_{0.29}$ \\
	& Cosine + L2	 & 0.628$_{0.24}$ & 0.119$_{0.03}$ & 0.816$_{0.36}$ & 0.602$_{0.46}$ \\
	& Cosine	 & 0.619$_{0.18}$ & 0.630$_{0.40}$ & 1.049$_{0.25}$ & 0.920$_{0.33}$ \\
	& L2	 & 0.629$_{0.32}$ & 0.613$_{0.40}$ & 0.869$_{0.43}$ & 0.860$_{0.36}$ \\
	& L1	 & 0.471$_{0.08}$ & 0.095$_{0.06}$ & \textbf{0.278$_{0.20}$} & \textbf{0.244$_{0.27}$} \\
\midrule
4	& Cosine + L1	 & 0.647$_{0.11}$ & 0.261$_{0.17}$ & 0.825$_{0.35}$ & \textbf{0.516$_{0.40}$} \\
	& Cosine + L2	 & 0.715$_{0.16}$ & 0.901$_{0.20}$ & 1.231$_{0.12}$ & 0.881$_{0.25}$ \\
	& Cosine	 & 0.738$_{0.18}$ & 0.818$_{0.33}$ & 1.132$_{0.34}$ & 0.933$_{0.43}$ \\
	& L2	 & 0.742$_{0.16}$ & 0.811$_{0.31}$ & 1.264$_{0.16}$ & 1.040$_{0.26}$ \\
	& L1	 & \textbf{0.554$_{0.11}$} & \textbf{0.131$_{0.10}$} & \textbf{0.777$_{0.34}$} & 0.539$_{0.45}$ \\
\bottomrule
\end{tabular}}
\end{table}

\section{Evaluation}
\label{sec:05_evaluation}
We first detail the experimental setup, followed by the attack accuracy of \algo, and an ablation study.
\subsection{Experimental setup}
\label{sec:experiment_setup}

\subsubsection{Dataset} For the datasets we have selected a total of 
four datasets. Three electricity datasets: Electricity (370)\footnote{\url{https://archive.ics.uci.edu/dataset/321/electricityloaddiagrams20112014}}, London Smartmeter\footnote{\url{https://zenodo.org/records/4656091}}, and a proprietary dataset. We have also opted for an additional dataset about air quality: KDDCup 2018 (Air Quality)\footnote{\url{https://zenodo.org/records/4656756}}. Even though the air quality is not a privacy-sensitive dataset, it evaluates the method's ability to generalize. The energy datasets are sampled with 1 observation day and 1 future day to forecast. The KDDCup dataset has 5 observation days and 2 future days. The sample description of the datasets is summarized in Table~\ref{tab:datasets}. All datasets are normalized using the min-max method bounding it between 0 and 1.

\subsubsection{Data Pre-processing}
\label{sec:experiment_dataset_creation}
We employ the rolling window method to preprocess the data, i.e., moving a window of fixed-width steps through the dataset, advancing by steps with each iteration. The data within this window is divided into observation and target segments $\mathbf{S}_{obs}$ and $\mathbf{S}_{tar}$ according to an observation length $H$ and future length $F$. The result of the rolling window process is a dataset sample pairs of observations $\mathbf{S}_{obs}$ and targets $\mathbf{S}_{tar}$. The dataset-specific settings are described in Table~\ref{tab:datasets}. The datasets are divided into training, validation, and test sets with ratios of 64\%, 16\%, and 20\%, respectively, by recursively applying a 20\% split ratio. For the auxiliary dataset $\mathcal{D}_{aux}$, the validation set is used.

\begin{figure}[]
    \centering
    \resizebox{\linewidth}{!}{
        \includegraphics[]{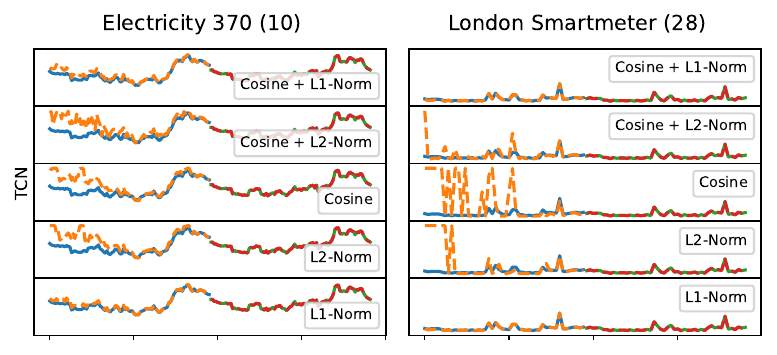}
    }
    \caption{\algo, reconstruction comparison on the TCN model architecture with different gradient distance functions and without regularization.}
    \label{fig:ts-inverse-gradient_distance-reconstructions}
\end{figure}

\subsubsection{Baselines}
The baselines that have been evaluated are the DLG-LBFGS~\cite{zhu_deep_2019}, DLG-Adam~\cite{zhu_deep_2019}, InvG~\cite{geiping_inverting_2020}, DIA~\cite{scheliga_dropout_2022}, and LTI~\cite{wu_learning_2023}. 
The implementation and configuration details are consistent with their open-sourced repositories. Where LTI is trained for 250 epochs and the other methods attack with 5000 iterations.

\subsubsection{Network Architectures}
\label{sec:06_exper_model_architectures}
The FCN model has 3 FC layers with sigmoid acivation functions, the CNN model is based on LeNet with 1D Conv layers. The TCN architecture has a kernel size of 6, dilation factor of 2 and the number of levels is adjusted such that the receptive field is encompassing the observation sequence. The GRU-2-FCN and GRU-2-GRU both have the standard GRU implementation. All models apart from the GRU-2-GRU have a FC head that is outputting the final values.  All hidden sizes of the architectures are 64.

The gradient inversion model $f_{inv}$ has two modules for outputting observations and targets, each containing two residual blocks with hidden sizes of 768 and 512, respectively. Each residual block consists of fully connected layers, ReLU activations, batch normalization, and dropout for regularization, with an optional adaptation layer ensuring matching dimensions for identity mapping. Additionally, each module includes a fully connected layer that outputs the quantiles for its observation or target reconstruction. The model is trained for 75 epochs and the quantile levels are $\{0.1, 0.3, 0.7, 0.9\}$.

\subsubsection{Evaluation Metrics}
To measure the reconstruction quality the Symmetrical Mean Absolute Percentage Error (sMAPE $\downarrow$) is used, which is bounded between 0.0 and 2.0, and allows for better comparisons between datasets. This normalization is achieved by scaling the absolute difference between the actual values ($\mathbf{S}$) and the predicted values ($\hat{\mathbf{S}}$) by the sum of their absolute values. The formula for sMAPE is given by:
\begin{equation*}
\text{sMAPE} = \frac{1}{N} \sum_{i=1}^{N} \frac{2 \left| \mathbf{S}_i - \hat{\mathbf{S}}_i \right|}{\left| \mathbf{S}_i \right| + \left| \hat{\mathbf{S}}_i \right|}
\end{equation*}
where $N$ are the number of elements in sequence $\mathbf{S}$, and $|\cdot|$ is the absolute value function. When using other metrics, such as MSE and MAE, the ranks do not change.

\subsection{Results Overview for \algo}
\begin{figure}[]
    \centering
    \resizebox{\linewidth}{!}{
        \includegraphics[]{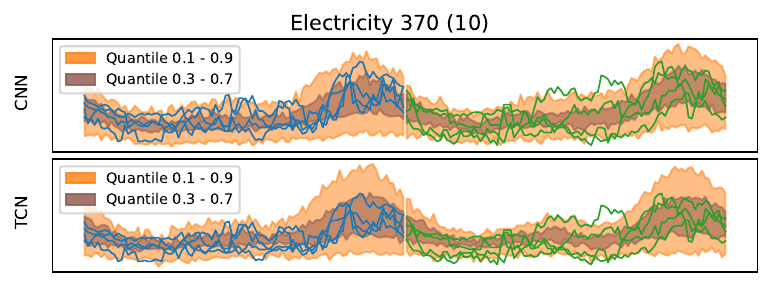}
    }
    \caption{\algo quantile predictions that function as prior in the learned regularization on Electricity 370 dataset for the CNN and TCN model architecture and $\mathcal{B} = 4$.}
    \label{fig:ts-inverse-quantile-reconstructions}
\end{figure}

The experimental results demonstrate that \algo outperforms the baseline methods in TS reconstruction across multiple datasets and architectures as shown in Table~\ref{tab:big_comparison}. Specifically, \algo achieves the lowest sMAPE values for both observation and target sequences. Figures~\ref{fig:ts-inverse_reconstructions} and~\ref{fig:ts-inverse-quantile-reconstructions} illustrate how \algo captures underlying patterns and quantile bounds in the reconstructed sequences. The subsequent experiments investigate the impact of different gradient distance functions (Table~\ref{tab:ts-inverse_tcn_gradient_loss_comparison}) and assess the effectiveness of various regularization techniques, including quantile bounds, periodicity, and trend regularization (Tables~\ref{tab:learned_prior_regularization}, \ref{tab:04_03_periodicity_regularization_terms}, and \ref{tab:04_03_trend_regularization_terms}). The combined effects of the various regularization techniques are displayed in Table~\ref{tab:ts-inverse-combination-regularization} and Figure~\ref{fig:ts-inverse-combination-regularization-reconstructions}. These studies highlight the robustness and versatility of \algo in time series inversion tasks across different architectures and settings.

\subsection*{Effectiveness of the Proposed Gradient Distance Function}
Table~\ref{tab:ts-inverse_tcn_gradient_loss_comparison} presents the sMAPE results for different gradient distance functions without additional regularization, focusing on capturing the effects of the gradient distances alone. The gradient distance functions tested include Cosine + L1-Norm, Cosine + L2-Norm, Cosine, L2, and L1. The results indicate that distance functions incorporating the L1-Norm perform better with lower sMAPE results compared to other functions. This trend is consistent across various datasets and is particularly evident in the TCN architecture, as shown in Figure~\ref{fig:ts-inverse-gradient_distance-reconstructions}, which displays individual reconstructions for $\mathcal{B} = 1$. For the baseline architectures (FCN and CNN), the differences in sMAPE results among the distance functions are minimal.

\begin{table*}[]
\caption{\algo results measured in sMAPE (↓) for the Electricity 370 dataset using quantile bounds regularization on observations (\(\lambda_Q^{obs}\)) and targets (\(\lambda_Q^{tar}\)). Results are shown for different hyperparameter values with the best performances in bold. Experiments are run with seeds 10 and 43, and the standard deviations are provided as subscripts.}
\label{tab:learned_prior_regularization}
\centering
\resizebox{0.75\linewidth}{!}{
\begin{tabular}{l|cc|cc|cc|cc}
\toprule
      $\lambda_Q^{obs}$                 & \multicolumn{2}{c}{0} & \multicolumn{2}{c}{0.1} & \multicolumn{2}{c}{0.5} & \multicolumn{2}{c}{1} \\
\midrule
$\lambda_Q^{tar}$ & Observation & Target           & Observation & Target           & Observation & Target           & Observation & Target \\
\midrule
 0	 & 0.515$_{0.11}$ & 0.068$_{0.05}$ & 0.504$_{0.12}$ & 0.185$_{0.11}$ & 0.207$_{0.08}$ & 0.061$_{0.06}$ & 0.203$_{0.05}$ & 0.060$_{0.03}$ \\
	0.1	 & 0.567$_{0.17}$ & 0.048$_{0.01}$ & 0.449$_{0.14}$ & 0.105$_{0.05}$ & 0.284$_{0.02}$ & 0.237$_{0.11}$ & \textbf{0.144$_{0.03}$} & \textbf{0.024$_{0.02}$} \\
	0.5	 & 0.568$_{0.18}$ & 0.067$_{0.02}$ & 0.368$_{0.21}$ & 0.026$_{0.02}$ & 0.259$_{0.08}$ & 0.112$_{0.08}$ & 0.207$_{0.04}$ & 0.131$_{0.06}$ \\
	1	 & 0.514$_{0.07}$ & 0.053$_{0.03}$ & 0.522$_{0.22}$ & 0.342$_{0.26}$ & 0.226$_{0.12}$ & 0.040$_{0.03}$ & 0.196$_{0.04}$ & 0.077$_{0.03}$ \\
\bottomrule
\end{tabular}}
\vspace{-5pt}
\end{table*}

\subsection*{Effectiveness of Regularization for Batch Reconstruction}
The effects of the hyperparameters for the quantile bounds regularization are shown in Table~\ref{tab:learned_prior_regularization}. The hyperparameters for the quantile bounds regularization are optimized through a grid search over the values $\lambda_Q^{obs} \in \{0, 0.1, 0.5, 1 \}$ and $\lambda_Q^{tar} \in \{0, 0.1, 0.5, 1 \}$. The results indicate that increasing the regularization term for observations leads to a decrease in the sMAPE metric, whereas the effect is less apparent for target sequences. From these results the optimal terms are found to be $\lambda_Q^{obs}=1$ and $\lambda_Q^{tar}=0.1$ for the Electricity 370 dataset.

\subsection*{Effectiveness of Periodicity Regularization}
\begin{figure}[]
    \centering
    \resizebox{\linewidth}{!}{
    \includegraphics[]{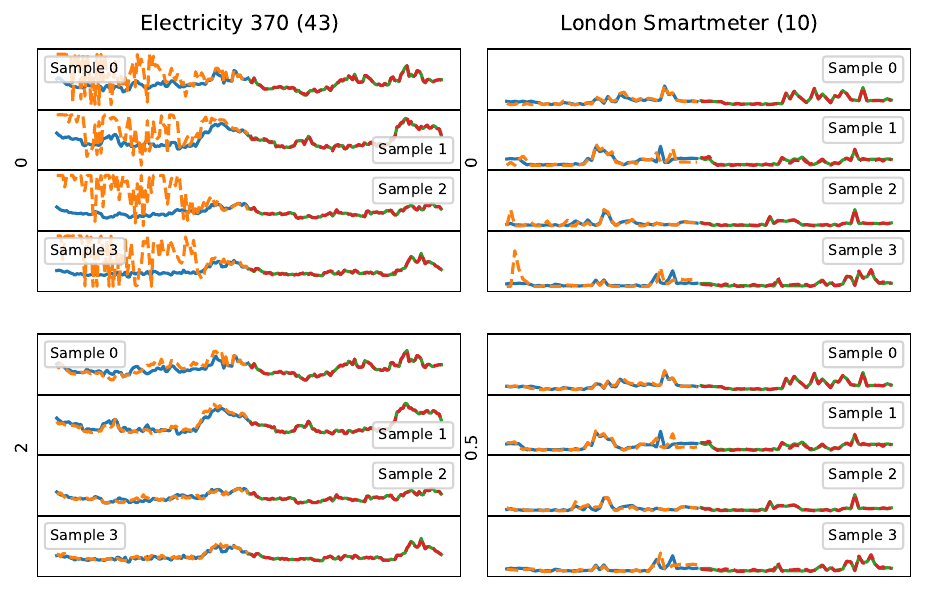}
    }
    \caption{\algo: The best periodicity regularization $\lambda_\text{P}$ compared to no regularization. Rows are $\lambda_\text{P}$ and Columns the dataset. Model: TCN, Datasets: Electricity 370, London Smartmeter, Batch size $\mathcal{B} = 4$, (seeds 10 and 43).}
    \label{fig:04_03_periodicity_regularization_specific}
\end{figure}

The impact of periodicity regularization on reconstruction performance is detailed in Table~\ref{tab:04_03_periodicity_regularization_terms}. We assessed various values of the periodicity regularization term $\lambda_{\text{P}} \in \{0.1, 0.5, 1, 2\}$ across three datasets: Electricity 370, London Smartmeter, and Proprietary. The results demonstrate that incorporating periodicity regularization significantly enhances the reconstruction quality, particularly for the Electricity 370 dataset. Specifically, setting $\lambda_{\text{P}} = 2$ yields the lowest sMAPE scores for both observations and targets, indicating superior reconstruction accuracy. This improvement is visually corroborated in Figure~\ref{fig:04_03_periodicity_regularization_specific}, which shows that the reconstructed sequences more closely align with the original data when periodicity regularization is applied. Notably, the optimal value of $\lambda_{\text{P}}$ varies across datasets; for the London Smartmeter dataset, the best performance is achieved at $\lambda_{\text{P}} = 0.5$, while for the Proprietary dataset, $\lambda_{\text{P}} = 1$ yields the most favorable results. These findings suggest that the effectiveness of periodicity regularization is dataset-dependent, and appropriate tuning of $\lambda_{\text{P}}$ is essential for maximizing reconstruction performance.

\begin{table}[]
\caption{\algo results measured in sMAPE (↓) for periodicity regularization with different $\lambda_{\text{P}}$ values using the TCN model. The results are shown for three datasets: Electricity 370, London Smartmeter, and Proprietary. The best performances are highlighted in bold. Experiments are conducted with a batch size of $\mathcal{B}=4$ using seeds 10 and 43, with standard deviations provided as subscripts.}
\label{tab:04_03_periodicity_regularization_terms}
\centering
\resizebox{\linewidth}{!}{
\begin{tabular}{l|cc|cc|cc}
\toprule
      Dataset                 & \multicolumn{2}{c}{Elec. 370} & \multicolumn{2}{c}{London Smart.} & \multicolumn{2}{c}{Proprietary} \\
\midrule
$\lambda_{\text{P}}$ & Observation & Target           & Observation & Target           & Observation & Target \\
\midrule
0	 & 0.515$_{0.11}$ & 0.068$_{0.05}$ & 0.603$_{0.29}$ & 0.250$_{0.22}$ & 0.956$_{0.12}$ & 0.197$_{0.03}$ \\
	0.1	 & 0.400$_{0.04}$ & 0.146$_{0.12}$ & 0.838$_{0.19}$ & 0.620$_{0.22}$ & 0.670$_{0.02}$ & 0.369$_{0.07}$ \\
	0.5	 & 0.306$_{0.16}$ & 0.219$_{0.18}$ & \textbf{0.467$_{0.18}$} & \textbf{0.250$_{0.21}$} & 0.396$_{0.02}$ & 0.214$_{0.03}$ \\
	1	 & 0.355$_{0.16}$ & 0.375$_{0.13}$ & 0.714$_{0.24}$ & 0.538$_{0.43}$ & \textbf{0.372$_{0.01}$} & \textbf{0.167$_{0.07}$} \\
	2	 & \textbf{0.163$_{0.05}$} & \textbf{0.063$_{0.05}$} & 0.658$_{0.06}$ & 0.289$_{0.14}$ & 0.452$_{0.02}$ & 0.193$_{0.01}$ \\
\bottomrule
\end{tabular}}
\vspace{-10pt}
\end{table}

\subsection*{Effectiveness of Trend Regularization}
\begin{figure}[]
    \centering
    \resizebox{\linewidth}{!}{
    \includegraphics[]{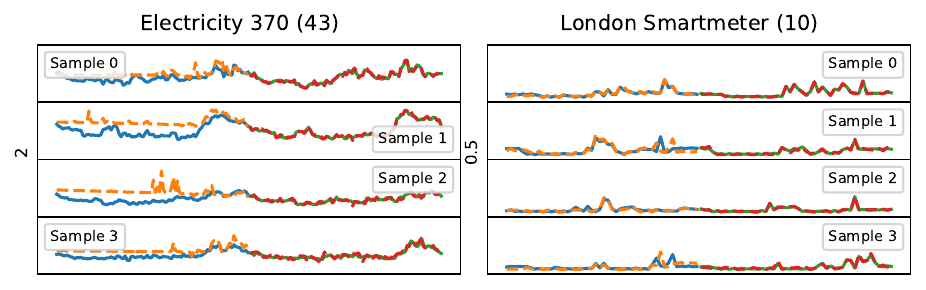}
    }
    \caption{\algo: The best trend regularization $\lambda_{T}$. Model: TCN, Datasets: Electricity 370, London Smartmeter, and Proprietary, Batch size $\mathcal{B}=4$, (seeds 10 and 43)}
    \label{fig:04_03_trend_regularization_specific}
\end{figure}

We further explored the influence of trend regularization on the reconstruction results, as presented in Table~\ref{tab:04_03_trend_regularization_terms}. By varying the trend regularization parameter $\lambda_{\text{T}} \in \{0.1, 0.5, 1, 2\}$, we evaluated its effect on the same three datasets. The results indicate that trend regularization enhances the model's ability to capture underlying trends in the data, leading to improved reconstruction accuracy. For the Electricity 370 dataset, the lowest sMAPE scores for observations are obtained at $\lambda_{\text{T}} = 2$, demonstrating the positive impact of stronger trend regularization. Figure~\ref{fig:04_03_trend_regularization_specific} visually illustrates this improvement, showing that the reconstructed sequences better reflect the original data's trend when the regularization is applied. Similar trends are observed for the London Smartmeter and Proprietary datasets, with optimal performances at $\lambda_{\text{T}} = 0.5$ and $\lambda_{\text{T}} = 1$, respectively. These results underscore the importance of incorporating trend regularization in the reconstruction process, especially for datasets where trend components are prominent. Overall, careful tuning of $\lambda_{\text{T}}$ is crucial to effectively leverage trend information and enhance reconstruction performance across different datasets.

\begin{table}[]
\caption{\algo results measured in sMAPE (↓) for trend regularization with different $\lambda_{\text{T}}$ values using the TCN model. The results are shown for three datasets: Electricity 370, London Smartmeter, and Proprietary. The best performances are highlighted in bold. Experiments are conducted with a batch size of $\mathcal{B}=4$ using seeds 10 and 43, with standard deviations provided as subscripts.}
\label{tab:04_03_trend_regularization_terms}
\centering
\resizebox{\linewidth}{!}{
\begin{tabular}{l|cc|cc|cc}
\toprule
      Dataset                 & \multicolumn{2}{c}{Elec. 370} & \multicolumn{2}{c}{London Smart.} & \multicolumn{2}{c}{Proprietary} \\
\midrule
$\lambda_{\text{T}}$ & Observation & Target           & Observation & Target           & Observation & Target \\
\midrule
 0	 & 0.515$_{0.11}$ & 0.068$_{0.05}$ & 0.603$_{0.29}$ & 0.250$_{0.22}$ & 0.956$_{0.12}$ & 0.197$_{0.03}$ \\
	0.1	 & 0.505$_{0.15}$ & \textbf{0.054$_{0.02}$} & 0.853$_{0.08}$ & 0.416$_{0.06}$ & 0.955$_{0.02}$ & 0.443$_{0.27}$ \\
	0.5	 & 0.392$_{0.08}$ & 0.152$_{0.12}$ & \textbf{0.495$_{0.21}$} & \textbf{0.247$_{0.23}$} & 0.736$_{0.10}$ & 0.482$_{0.30}$ \\
	1	 & 0.419$_{0.17}$ & 0.324$_{0.21}$ & 0.685$_{0.04}$ & 0.568$_{0.14}$ & \textbf{0.475$_{0.12}$} & \textbf{0.145$_{0.02}$} \\
	2	 & \textbf{0.281$_{0.05}$} & 0.124$_{0.06}$ & 0.633$_{0.01}$ & 0.250$_{0.21}$ & 0.515$_{0.06}$ & 0.242$_{0.07}$ \\
\bottomrule
\end{tabular}}
\vspace{20pt}
\end{table}

\subsection*{Effectiveness of Combined Regularization}
\begin{figure}[]
    \centering
    \begin{subfigure}[b]{0.45\linewidth}
        \centering
        \includegraphics[width=\linewidth]{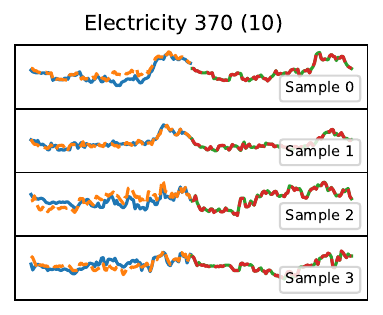}
    \end{subfigure}
    \begin{subfigure}[b]{0.45\linewidth}
        \centering
        \includegraphics[width=\linewidth]{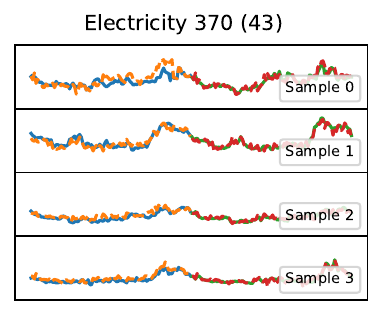}
    \end{subfigure}
    \caption{\algo reconstructions with all regularization applied for the TCN model architecture and $\mathcal{B} = 4$.}
    \label{fig:ts-inverse-combination-regularization-reconstructions}
\vspace{-10pt}
\end{figure}

We investigated the combined effects of periodicity and trend regularization on reconstruction performance, as summarized in Table~\ref{tab:ts-inverse-combination-regularization}. The parameters for both regularization terms, $\lambda_{P}$ and $\lambda_{T}$, were tested with values $\{0.1, 0.5, 1\}$. The results indicate that the integration of periodicity and trend regularization significantly enhances reconstruction accuracy. Specifically, the best performance was achieved when both regularization methods were applied together, especially in the presence of quantile bounds. Figure~\ref{fig:ts-inverse-combination-regularization-reconstructions} visually demonstrates the improvements in reconstructed sequences with combined regularization, highlighting the advantages of this approach in capturing complex patterns in the data. The findings underscore the necessity of tuning both $\lambda_{P}$ and $\lambda_{T}$ to optimize reconstruction across different datasets.

\begin{table*}[]
\caption{\algo results measured in sMAPE (↓) for the Electricity 370 dataset using periodicity and trend regularization ($\mathcal{R}_P + \mathcal{R}_T$) with and without quantile bounds regularization ($\mathcal{R}_Q$). Results are shown for different regularization parameter values ($\lambda_P$ and $\lambda_T$) with the best performances in bold. Experiments are ran with seeds 10 and 43 and the standard deviations are provided as subscripts.}
\label{tab:ts-inverse-combination-regularization}
\centering
\resizebox{0.8\linewidth}{!}{
\begin{tabular}{l|l|cc|cc|cc}
\toprule
      & $\lambda_{P}$                 & \multicolumn{2}{c}{0.5} & \multicolumn{2}{c}{1} & \multicolumn{2}{c}{2} \\
\midrule
$\lambda_{T}$ & Regularization & Observation & Target           & Observation & Target           & Observation & Target \\
\midrule
0.5	& $\mathcal{R_P} + \mathcal{R}_T$ & \textbf{0.154$_{0.03}$} & \textbf{0.118$_{0.10}$} & 0.155$_{0.05}$ & 0.086$_{0.06}$ & 0.350$_{0.14}$ & 0.302$_{0.18}$ \\
	& $\mathcal{R_P} + \mathcal{R}_T + \mathcal{R}_Q$	 & 0.273$_{0.12}$ & 0.287$_{0.19}$ & \textbf{0.124$_{0.00}$} & \textbf{0.039$_{0.03}$} & 0.174$_{0.02}$ & 0.199$_{0.01}$ \\
\midrule
1	& $\mathcal{R_P} + \mathcal{R}_T$	 & 0.197$_{0.04}$ & 0.075$_{0.02}$ & 0.350$_{0.20}$ & 0.330$_{0.29}$ & 0.162$_{0.01}$ & 0.081$_{0.01}$ \\
	& $\mathcal{R_P} + \mathcal{R}_T + \mathcal{R}_Q$	 & 0.161$_{0.01}$ & 0.059$_{0.02}$ & 0.141$_{0.01}$ & 0.080$_{0.05}$ & 0.158$_{0.01}$ & 0.098$_{0.03}$ \\
\midrule
2	& $\mathcal{R_P} + \mathcal{R}_T$	 & 0.217$_{0.02}$ & 0.058$_{0.01}$ & 0.184$_{0.02}$ & 0.102$_{0.01}$ & 0.314$_{0.15}$ & 0.254$_{0.21}$ \\
	& $\mathcal{R_P} + \mathcal{R}_T + \mathcal{R}_Q$	 & 0.184$_{0.02}$ & 0.082$_{0.02}$ & 0.132$_{0.02}$ & 0.051$_{0.04}$ & 0.124$_{0.03}$ & 0.114$_{0.09}$ \\
\bottomrule
\end{tabular}}
\end{table*}

\subsection{Defenses}
In addition to conventional FL defense strategies, such as pruning \cite{alham_sparse_2017}, differential privacy \cite{abadi_deep_2016}, and sign compression \cite{jeremy_signsgd_2018}, we explore the defense capabilities of GRU-based forecasting models.

Foremost, Table~\ref{tab:ts-defense} illustrates the outcomes of these three defense mechanisms. The findings suggest that the LTI attack shows greater performance in attacking the defense mechanisms than the optimization-based attacks. We hypothesis that this is due to the fact that the Inversion model is trained with gradients on which the defense mechanisms are applied, and thus the model could potentially learn the defense characteristics and counter this.

\begin{figure}[]
    \centering
    \resizebox{\linewidth}{!}{
        \includegraphics[]{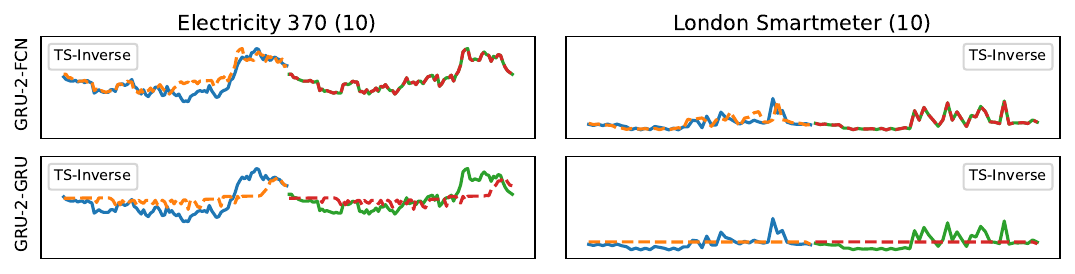}
    }
    \caption{\algo reconstructions on Electricity 370 and London Smartmeter datasets for the GRU-2-FCN and GRU-2-GRU architectures $\mathcal{B} = 1$ with $\lambda_P = 1$, $\lambda_T = 0.5$, $\lambda_Q^{obs} = 1$, and $\lambda_Q^{tar} = 0.1$.}
    \label{fig:ts-inverse-gru-reconstructions}
\vspace{-5pt}
\end{figure}

Furthermore, we evaluate using GRU in the forecasting model to counter \algo. GRU-based forecasting models are more robust against attacks due to the many-to-one mapping caused by the recurrent design. We attack the GRU-2-FCN and GRU-2-GRU models, with reconstruction results shown in Figure~\ref{fig:ts-inverse-gru-reconstructions}. The results show that the targets associated with the FCN part remain vulnerable. We hypothesize that regularization causes the values to transfer to the observations, as the observations of the GRU-2-FCN architecture closely resemble the targets. This is backed by the results of employing a GRU-2-GRU architecture, which prevents detailed reconstruction of the observations and targets. However, the attack does seem to be able to extract the trend from the gradients. Overall, we the GRU-based architectures enhance the robustness against privacy attacks because of the many-to-one mapping design.
    
\begin{table}[]
\caption{Baselines and \algo results measured in sMAPE (↓) for the Electricity 370 dataset and FCN model under various defense strategies. Results are presented for Observations and Targets, with the best performances highlighted in bold. Experiments were conducted with seeds 10, 43 and 28, and the standard deviations are indicated as subscripts.}
\label{tab:ts-defense}
\centering
\resizebox{0.75\linewidth}{!}{
\begin{tabular}{l|l|cc}
\toprule
      & Dataset                 & \multicolumn{2}{c}{Electricity 370} \\
\midrule
Defense & Attack Method & Observation & Target \\
\midrule
none	& DLG-LBFGS	 & 0.033$_{0.00}$ & 3.1e-04$_{0.00}$ \\
	& DLG-Adam	 & \textbf{1.5e-06$_{0.00}$} & 8.6e-06$_{0.00}$ \\
	& InvG	 & 4.4e-06$_{0.00}$ & 0.279$_{0.06}$ \\
	& DIA	 & 0.004$_{0.00}$ & 0.867$_{0.09}$ \\
	& LTI	 & 0.069$_{0.01}$ & 0.042$_{0.01}$ \\
	& \algo	 & 1.1e-05$_{0.00}$ & \textbf{8.5e-08$_{0.00}$} \\
\midrule
gauss	& DLG-LBFGS	 & 1.998$_{0.00}$ & 1.417$_{0.05}$ \\
	& DLG-Adam	 & 1.910$_{0.01}$ & 1.457$_{0.04}$ \\
	& InvG	 & 1.810$_{0.02}$ & 1.213$_{0.07}$ \\
	& DIA	 & 1.805$_{0.04}$ & 1.377$_{0.04}$ \\
	& LTI	 & \textbf{0.222$_{0.02}$} & \textbf{0.231$_{0.07}$} \\
	& \algo	 & 0.593$_{0.04}$ & 1.179$_{0.13}$ \\
\midrule
prune	& DLG-LBFGS	 & 0.034$_{0.00}$ & 0.009$_{0.00}$ \\
	& DLG-Adam	 & \textbf{0.003$_{0.00}$} & 0.009$_{0.00}$ \\
	& InvG	 & 0.003$_{0.00}$ & 0.284$_{0.05}$ \\
	& DIA	 & 0.003$_{0.00}$ & 0.861$_{0.09}$ \\
	& LTI	 & 0.071$_{0.01}$ & 0.047$_{0.01}$ \\
	& \algo	 & 0.011$_{0.01}$ & \textbf{0.009$_{0.00}$} \\
\midrule
sign	& DLG-LBFGS	 & 1.963$_{0.03}$ & 1.989$_{0.00}$ \\
	& DLG-Adam	 & 1.929$_{0.01}$ & 1.951$_{0.02}$ \\
	& InvG	 & 1.672$_{0.01}$ & 0.816$_{0.12}$ \\
	& DIA	 & 1.672$_{0.01}$ & 1.261$_{0.13}$ \\
	& LTI	 & \textbf{0.324$_{0.07}$} & \textbf{0.184$_{0.06}$} \\
	& \algo	 & 1.049$_{0.06}$ & 1.230$_{0.14}$ \\
\bottomrule
\end{tabular}}
\vspace{-10pt}
\end{table}

\section{Related Studies}
\begin{table*}[]
\caption{Comparison of related gradient inversion attacks and \algo. The space indicates whether the attack optimizes the inputs ($\mathbf{x}$), outputs ($\mathbf{y}$), latent ($\mathbf{z}$), or generative model parameters ($G_\theta$). The regularization terms mentioned are Total Variation, Clipping, Scaling, Dropout Masks, Batch and Group Normalization, Softmax Relaxation, Periodicity, Trend and Quantile Bounds. The FL Training Loss indicates with which loss the global model is trained.}
\label{tab:related_studies}
\centering
\resizebox{\linewidth}{!}{
    \begin{tabular}{l|l|l|l|l|l|l}
    \toprule
    \textbf{Approach} & \textbf{Attack Name / Source} & \textbf{Type of attack} & \textbf{Space} & \textbf{Gradient Distance} & \textbf{Regularization} & \textbf{FL Training Loss} \\ 
    \midrule
    Pure 
    & DLG \cite{zhu_deep_2019} & Optimization & $(\mathbf{x}, \mathbf{y})$ & L2 & - & Cross Entropy \\ 
    & iDLG \cite{zhao_idlg_2020} & Optimization & $(\mathbf{x})$ & L2 & - & Cross Entropy \\ 
    & Inverting Gradients (InvG) \cite{geiping_inverting_2020} & Optimization & $(\mathbf{x})$ & Cosine & TV & Cross Entropy \\ 
    & SAPAG \cite{wang_sapag_2020} & Optimization & $(\mathbf{x}, \mathbf{y})$ & Weighted Gaussian Kernel & - & Cross Entropy \\ 
    & R-GAP \cite{zhu_r-gap_2021} & Recursive & $(\mathbf{x})$ & - & - & Cross Entropy \\ 
    & AGIC \cite{xu_agic_2022} & Optimization & $(\mathbf{x}, \mathbf{y})$ & Cosine & TV & Cross Entropy \\ 
    & DIA \cite{scheliga_dropout_2022} & Optimization & $(\mathbf{x})$ & Cosine & TV, Mask & Cross Entropy \\ 
    & General DL \cite{geng_towards_2022, geng_improved_2023} & Optimization & $(\mathbf{x})$ & L2 & TV, Clip, Scale & Cross Entropy \\ 
    & TAG \cite{deng_tag_2021} & Optimization & $(\mathbf{x}, \mathbf{y})$ & L2 + $\alpha$ L1 & - & Causal (Cross Ent.) \\ 
    & TabLeak \cite{vero_tableak_2023} & Optimization & $(\mathbf{x})$ & Cosine & Softmax Relaxation, Ensemble & Cross Entropy \\ 
    \midrule
    Informed & GradInversion \cite{yin_see_2021} & (Latent) Optimization & $(\mathbf{z}, \mathbf{x})$ & L2 & TV, BN \& Group & Cross Entropy \\ 
    & GIAS / GIML \cite{jeon_gradient_2021} & (Latent) Optimization & $(\mathbf{z}, G_\theta)$ & Cosine & TV, L2 & Cross Entropy \\ 
    & LTI \cite{wu_learning_2023} & Learning & $(\mathbf{x})$ & Permutation Invariant CE or L2 & - & - \\ 
    & \algo (this work) & Learning \& Optimization & $(\mathbf{x}, \mathbf{y})$ & L1 & Periodicity, Trend \& QB & MSE \\ 
    \bottomrule
    \end{tabular}
    }
\end{table*}

Existing work on inversion attacks can be categorized into two types: pure optimization-based and informed optimization-based as presented in Table \ref{tab:related_studies}. Both categories optimize dummy data such that their gradients match the observed gradients. The informed optimization-based attacks further leverage external knowledge (e.g., generative models) to assist the reconstruction process. 
The exploration of gradient inversion attacks reveals a rich and evolving landscape of techniques aimed at reconstructing training data from gradients. The foundational work "Deep Leakage from Gradients" (DLG)~\cite{zhu_deep_2019} set the stage by demonstrating the feasibility of these attacks using Euclidean distance minimization. The subsequent enhancement, "Improved Deep Leakage from Gradients" (iDLG)~\cite{zhao_idlg_2020}, advanced this by analytically reconstructing training labels, though it remained limited to a batch size of one.
"Inverting Gradients" (InvG)~\cite{geiping_inverting_2020} introduced cosine similarity and total variation regularization, improving the attack's robustness, particularly for image data. The "Self-Adaptive Privacy Attack from Gradients" (SAPAG)~\cite{wang_sapag_2020} tackled convergence issues arising from different weight initializations, further refining the optimization process with a Gaussian kernel-based distance function.
The "Recursive Gradient Attack on Privacy" (R-GAP)~\cite{zhu_r-gap_2021} deviated from optimization-based methods, offering a faster, deterministic solution by recursively solving linear equations, albeit limited to single-sample batches. "See through Gradients"~\cite{yin_see_2021} and "Gradient Inversion with Generative Image Prior"~\cite{jeon_gradient_2021} provided significant advancements in reconstruction quality and scalability to larger batches, leveraging auxiliary regularizations and generative models.
The "Gradient Attack on Transformer-based Language Models" (TAG)~\cite{deng_tag_2021} addressed the unique challenges of NLP models, using combined Euclidean and Manhattan distances for gradient comparison. The "Approximate Gradient Inversion Attack" (AGIC)~\cite{xu_agic_2022} efficiently approximated multiple gradient updates in federated averaging, enhancing practicality for real-world federated learning setups.
The "Dropout Inversion Attack" (DIA)~\cite{scheliga_dropout_2022} extended the attack surface by accounting for dropout layers, effectively neutralizing this common defense mechanism. "Improved Gradient Inversion Attacks and Defenses"~\cite{geng_improved_2023} proposed a comprehensive framework addressing both FedSGD and FedAvg strategies, introducing zero-shot batch label inference and auxiliary regularizations for realistic image restoration.
"Learning to Invert" (LTI)~\cite{wu_learning_2023} presented a shift by using a model trained on an auxiliary dataset to predict client samples from gradients, overcoming traditional defenses like differential privacy. "Tabular Data Leakage" (TabLeak)~\cite{vero_tableak_2023} innovatively tackled the complexities of reconstructing tabular data, employing softmax relaxation and pooled ensembling for improved accuracy and reliability.
\section{Conclusion}
This study presents the first empirical analysis and effective algorithm, \algo, of reconstructing TS data for federated TSF models. Our empirical analysis demonstrates that existing inversion methods, developed for image and text classification, are less effective for TSF due to the mismatch between gradient distance and the gradient magnitudes and TS-specific model architecture, e.g., TCN \& GRU. To address these challenges, we propose \algo, which leverages a gradient inversion model and TS-specific characteristics as regularization during reconstruction. Specifically, the proposed TS-regularization include periodicity and trend. Furthermore, the gradient inversion model is trained with an auxiliary dataset to map gradients to quantile bounds of observations and targets. These bounds are also used to regularize the reconstruction. We extensively evaluate \algo on four data sets against five baselines, showing a 2x-10x  improvement in reconstruction sMAPE and indicating the potential of GRU-based architecture in defending against inversion attacks.

\section*{Acknowledgments}
This research was partly funded by  the SNSF project, Priv-GSyn 200021E\_229204.

\printbibliography

\end{document}